\documentclass[letterpaper]{article} 
\usepackage{aaai24}  
\usepackage{times}  
\usepackage{helvet}  
\usepackage{courier}  
\usepackage[hyphens]{url}  
\usepackage{graphicx} 
\usepackage{natbib}  
\usepackage{caption} 
\usepackage{algorithm}
\usepackage{algorithmic}

\usepackage{newfloat}
\usepackage{listings}
\DeclareCaptionStyle{ruled}{labelfont=normalfont,labelsep=colon,strut=off} 
\floatstyle{ruled}
\newfloat{listing}{tb}{lst}{}
\floatname{listing}{Listing}
\usepackage{makecell}
\usepackage{amsmath}
\usepackage{amsfonts}
\usepackage{amssymb}
\usepackage{booktabs}
\usepackage{diagbox}
\usepackage[symbol]{footmisc}

\newcommand{\fref}[1]{Figure \ref{#1}}
\newcommand{\Sref}[1]{Section \ref{#1}}
\newcommand{\sref}[1]{Section \ref{#1}}

\newcommand{\tref}[1]{Table \ref{#1}}

\newcommand{\eref}[1]{Eq. \ref{#1}}

\newcommand{\dir}{\mathbf{d}}  
\newcommand{\pos}{\mathbf{x}}  
\newcommand{\pe}{\gamma{}}  
\newcommand{\ray}{\mathbf{r}}  
\newcommand{\C}{\mathbf{C}}  
\newcommand{\Chat}{\hat{\mathbf{C}}}  
\newcommand{\mlpparam}{\Theta}  
\newcommand{\density}{\sigma}  %
\newcommand{\radiance}{\mathbf{c}}  %
\newcommand{\offset}{\delta}  
\newcommand{\real}{\mathbb{R}}
\title{Sync-NeRF: Generalizing Dynamic NeRFs to Unsynchronized Videos}
\author{
    Seoha Kim\textsuperscript{\rm 1}\equalcontrib,
    Jeongmin Bae\textsuperscript{\rm 1}\equalcontrib,
    Youngsik Yun\textsuperscript{\rm 1}, \\
    Hahyun Lee\textsuperscript{\rm 2},
    Gun Bang\textsuperscript{\rm 2},
    Youngjung Uh\textsuperscript{\rm 1}\thanks{Corresponding author.}
}

\affiliations{

    \textsuperscript{\rm 1}Yonsei University\\
    \textsuperscript{\rm 2}Electronics and Telecommunications Research Institute \\
    \{hailey07, jaymin.bae, bbangsik, yj.uh\}@yonsei.ac.kr, \{hanilee, gbang\}@etri.re.kr
}

\usepackage{bibentry}

\begin{document}

\maketitle

\begin{abstract}
Recent advancements in 4D scene reconstruction using neural radiance fields (NeRF) have demonstrated the ability to represent dynamic scenes from multi-view videos. However, they fail to reconstruct the dynamic scenes and struggle to fit even the training views in \emph{unsynchronized} settings. It happens because they employ a single latent embedding for a frame while the multi-view images at the same frame were actually captured at different moments. To address this limitation, we introduce time offsets for individual unsynchronized videos and jointly optimize the offsets with NeRF. By design, our method is applicable for various baselines and improves them with large margins. Furthermore, finding the offsets naturally works as synchronizing the videos without manual effort. Experiments are conducted on the common Plenoptic Video Dataset and a newly built Unsynchronized Dynamic Blender Dataset to verify the performance of our method. Project page: \url{https://seoha-kim.github.io/sync-nerf}
\end{abstract}

\section{Introduction}

Neural radiance fields (NeRF, \citealt{mildenhall2021nerf}) aim to synthesize novel views of a scene when given its limited number of views. As NeRF is a function that receives 3D coordinates with viewing directions and produces color and density, adding time as input inherently generalizes it to design dynamic NeRFs for multi-view videos. Recent works have improved efficiency \cite{li2022neural,TiNeuVox,Wang_2022_CVPR,wang2022mixed, fridovich2023k} and streamability \cite{li2022streaming, song2023nerfplayer, attal2023hyperreel} of dynamic NeRFs.

\begin{figure}[tb!]
  \centering
  \includegraphics[width=1\linewidth]{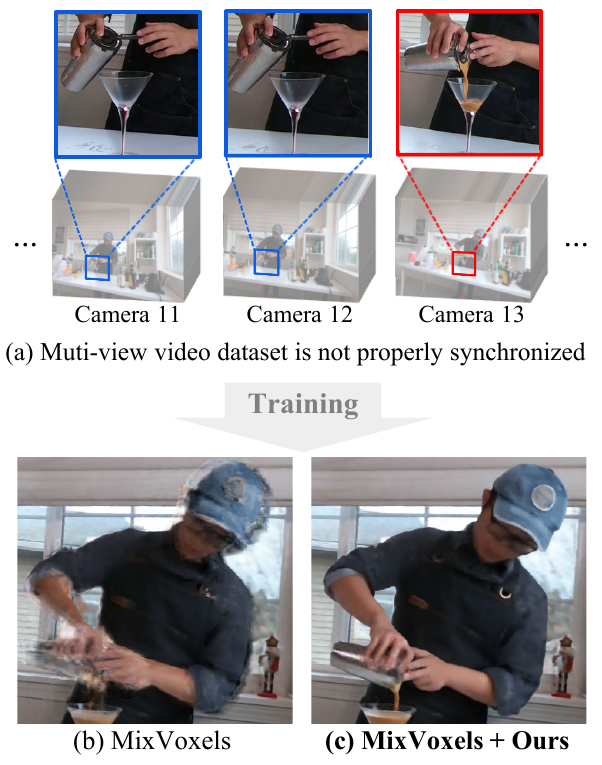}
  \caption{\textbf{Overview.} (a) The commonly used Plenoptic Video Dataset in 4D scene reconstruction contains an unsynchronized video. Image patches are all first frames. (b) If we include this view in the training set, baselines fail to reconstruct the motion around the unsynchronized viewpoint.
  (c) In the same settings, our method significantly outperforms.}
  \label{fig:teaser}
\end{figure}

Our research is motivated by the inaccurate video synchronization in a widely used multi-view dynamic dataset \cite{li2022neural}. Although typical synchronization approaches, such as timecode systems or audio peaks, usually work, these processes require an additional device, cannot be applied in noisy environments, or can be inaccurate \cite{li2022neural}. For example,  \fref{fig:teaser}a shows that the rightmost video is severely ahead of others in the temporal axis. While previous dynamic NeRFs show high-fidelity reconstruction by omitting the unsynchronized video, they fail to reconstruct this video even in the training view if it is included (\fref{fig:teaser}b). If we perturb the synchronization of the multi-view videos on purpose, all methods fail to reconstruct movements and produce severe artifacts and ghost effects.
 
\begin{figure*}[htb!]
  \centering
  \includegraphics[width=1\linewidth]{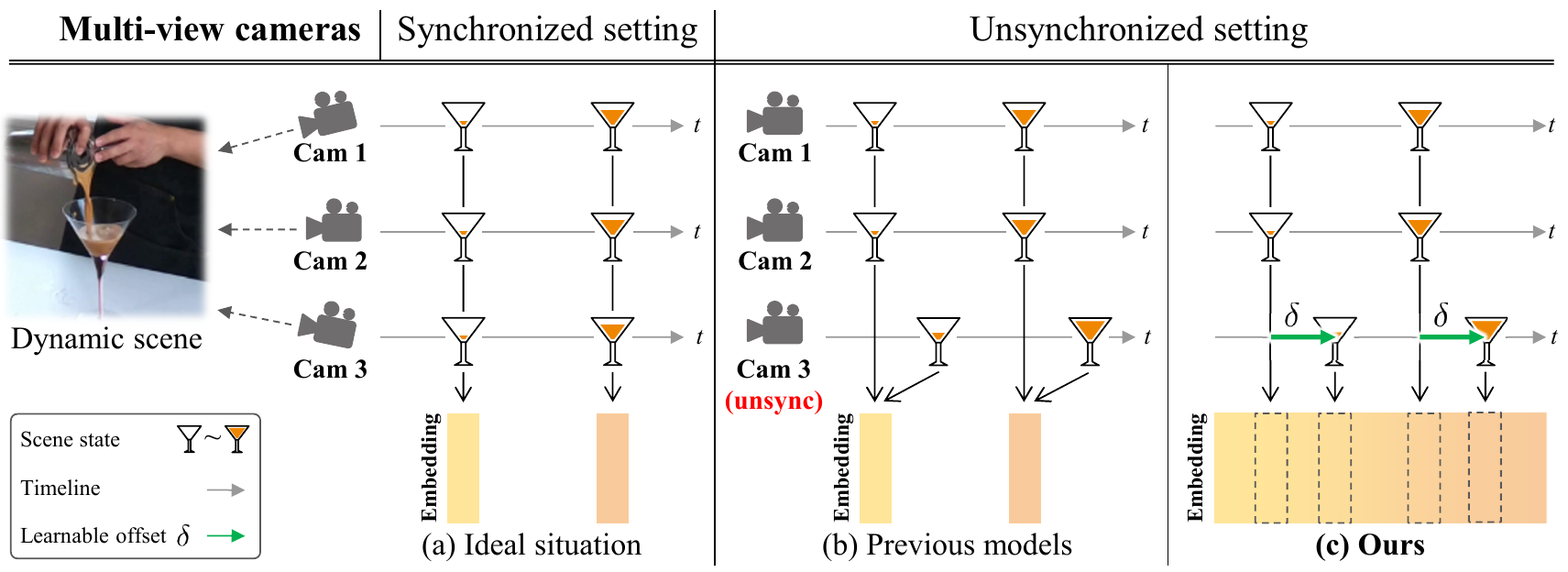}
  \caption{\textbf{Problem statement.} (a) Ideally, all multi-view images at a frame captures the same moment of a scene. Each frame is represented by a latent embedding. (b) Some frames are not synchronized. Previous methods suffer from the discrepancy between the latent embedding of the frame and the actual status of the scene. (c) Our method allows assigning correct temporal latent embeddings to videos captured with temporal gaps by introducing learnable time offsets $\offset$ for individual cameras.}
  \label{fig:offset}
\end{figure*}

The above drawbacks occur because existing dynamic NeRFs assume that the same frames in multi-view videos are captured at the same time (\fref{fig:offset}a), which is not always true. Even if the videos are assumed to be synchronized, there can be temporal mismatch within a frame. See Appendix A.2 for details. \fref{fig:offset}b illustrates the common problem: using the same temporal representation for different states of the scene. This problem worsens as the motions in the scene are faster, or as the deviation of time increases. In this paper, we introduce time offsets for individual videos and optimize them jointly with NeRF. It resolves the temporal gap between the observation and the target state caused by unsynchronization (\fref{fig:offset}c). Consequently, all training videos can be accurately reconstructed as shown in (\fref{fig:teaser}c). As optimizing the time offsets is equivalent to synchronizing the videos by shifting, we name our approach Sync-NeRF. We further design a continuous function that receives time and produces temporal representation to apply our method on dynamic NeRFs with discrete temporal representation \cite{wang2022mixed}. We inherit bilinear interpolation for dynamic NeRFs with spatiotemporal planes \cite{fridovich2023k}. 

In order to show multiple advantages of Sync-NeRF, we design new benchmark datasets by randomly unsynchronizing existing datasets and building unsynchronized synthetic scenes with ground truth time offsets.


\section{Related Work}
\paragraph{Dynamic NeRFs} have been evolving by introducing better representation for specific purposes. D-NeRF or HumanNeRF models deformation field to extend the static NeRF to dynamic domain \cite{pumarola2020d, zhao2022humannerf}. DyNeRF achieves complex temporal representation by implicitly assigning a latent vector to each frame \cite{li2022neural}. NeRFPlayer or MixVoxels improve the streamability of dynamic scenes by utilizing grid-based NeRF \cite{li2022neural, wang2022mixed}. K-Planes and HexPlane adopt planar factorization for extending to arbitrary dimensions, enabling the representation of dynamic scenes \cite{fridovich2023k,cao2023hexplane}. These works assume multiple synchronized video inputs but are limited to accurately synchronized multi-view videos. We tackle this limitation by introducing easily optimizable time offsets to the existing methods. Also, our approach can be seamlessly adapted to existing methods.

On the other hand, the methods for monocular video settings are relatively free from the synchronization \cite{park2021hypernerf, li2023dynibar}. Hence, We do not consider a monocular video setting. Nevertheless, we note that they rely on unnatural teleporting cameras \cite{gao2022dynamic}.

\paragraph{Per-frame Latent Embedding}
Some methods extend NeRF with multiple latent embeddings to represent multiple scenes or different states of a scene. NeRF-W \cite{martinbrualla2020nerfw} and Block-NeRF \cite{tancik2022block} use per-image appearance embeddings for different states of a scene to reconstruct a scene from unstructured image collections with appearance variations. In dynamic NeRFs, D-NeRF represents dynamic scenes using per-frame deformation embedding. Nerfies and HyperNeRF \cite{park2021nerfies, park2021hypernerf} apply both per-frame appearance embedding and per-frame deformation embedding. However, per-frame latent embedding approaches cannot represent a dynamic scene from unsynchronized multi-view videos. This is because they share a single latent embedding for multi-view images with the same frame index, even though these images are captured at different moments. Using our time offset method, existing dynamic NeRFs can represent dynamic scenes successfully in unsynchronized settings.

\paragraph{Joint Camera Calibration} 
NeRF$--$, BARF, and SCNeRF \cite{jeong2021self,lin2021barf,jeong2021self} optimize camera parameters and NeRF parameters jointly to eliminate the requirements of known camera parameters in the static setting. RoDyNeRF \cite{liu2023robust} optimizes dynamic NeRF jointly with camera parameters to tackle the failure of COLMAP in highly dynamic scenes. However, previous methods can not compensate for the inaccurate synchronization across multi-view videos. Traditional approaches for synchronization utilize timecode systems \cite{li2022neural} or audio peaks recorded simultaneously with videos \cite{shrstha2007synchronization}. Consequently, these methods require additional devices, and they may not produce accurate results or cannot be applied to videos with significant noise. We overcome these limitations by jointly training per-camera time offsets on top of the existing dynamic NeRFs.


\section{Method}
In this section, we introduce per-camera time offsets, which enable training dynamic NeRFs on the unsynchronized multi-view videos (\sref{sec:offset}). Subsequently, we describe an implicit function-based approach for models with per-frame temporal embeddings (\sref{sec:implicit}) and an interpolation-based approach for grid-based models (\sref{sec:grid}).


\subsection{Per-camera Time Offset with Dynamic NeRF}
\label{sec:offset}
NeRF learns to map a given 3D coordinates $\pos\in\real^3$ and viewing direction $\dir\in\real^3$ to RGB color $\radiance\in\real^3$ and volume density $\density\in\real$. To represent a dynamic scene with NeRF, it is common to modify NeRF as a time-dependent function by adding a time input $t\in\real$:

\begin{equation}
  \mathcal{F}_\mlpparam: (\pos,\dir,t)\rightarrow (\radiance,\density),
  \label{eq:base_func}
\end{equation}
where $\mlpparam$ parameterizes $\mathcal{F}$. Dynamic NeRFs typically employ the video frame index as $t$.
Then the model is trained to reconstruct multi-view videos by rendering each frame.

However when the multi-view videos are not synchronized, a single frame index may capture different moments of the scene across the different videos. As a result, the ground truth RGB images captured from different viewpoints at frame $t$ do not match each other, leading to suboptimal reconstruction of dynamic parts.

To this end, we introduce a learnable time offset $\offset_k$ for each of the $K$ training cameras: $t_k=t+\offset_{k}$. It allows the temporal axis of each video to be freely translated, rectifying potential temporal discrepancies across multi-view videos. The time-dependent function $\mathcal{F}_\mlpparam$ in \eref{eq:base_func} changes accordingly: 

\begin{equation}
  \mathcal{F}_\mlpparam: (\pos,\dir, t_k)\rightarrow (c,\sigma). 
  \label{eq:func_ours}
\end{equation}

We design the time offsets to be continuous rather than discrete frame indices. Further details are deferred to \sref{sec:implicit} and \ref{sec:grid}. The time offsets are jointly optimized with NeRF parameters by minimizing MSE between the ground truth RGB pixel $\C$ and the volume-rendered RGB pixel $\Chat$:

\begin{equation}
  \mathcal{L}_{\text{RGB}} = \sum_{k, \ray, t} \left\| \Chat(\ray, t+\offset_k) - \C_k(\ray, t) \right\|^2_2,
  \label{eq:render_our}
\end{equation}
where $k, \ray, t$ are the camera index, center ray, and time of each pixel in the training frames, respectively.
To calculate $\Chat$, we use the same numerical quadrature that approximates volume rendering integral as previous papers \cite{max1995optical,mildenhall2021nerf}. The time offsets resolve the disagreement across multi-view supervisions and significantly improve the reconstruction quality, especially on the dynamic parts.

\begin{figure}[tb!]
  \centering
  \includegraphics[width=1\linewidth]{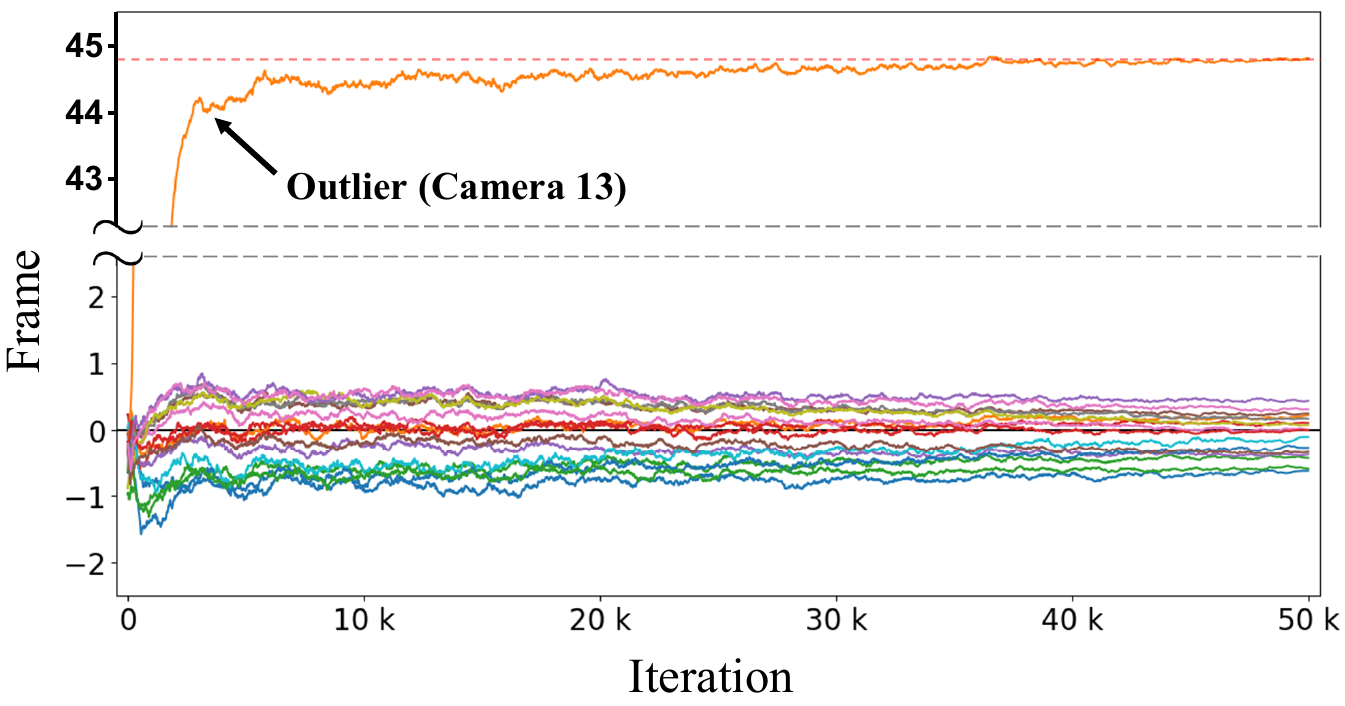}
  \caption{\textbf{Learning curve of time offsets.} We show camera offsets in \texttt{coffee\_martini} scene along the training iterations of Sync-MixVoxels. Our method successfully finds the offset of the outlier camera.
  }
\label{fig:offset_hist}
\end{figure}

\begin{figure}[tb!]
  \centering
  \includegraphics[width=1\linewidth]{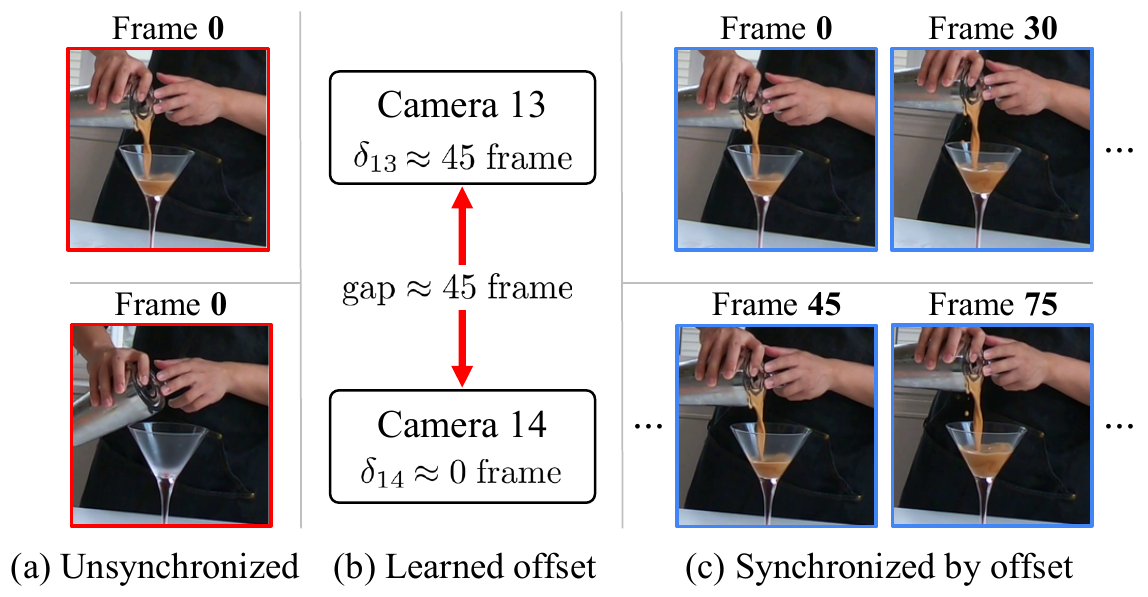}
  \caption{\textbf{Synchronization with time offsets.} (a) For given unsynchronized videos, (b) our method finds the time offsets $\offset$ which are equivalent to (c) automatically synchronizing the videos.
  }
\label{fig:offset_qual}
\end{figure}

\fref{fig:offset_hist} is an exemplar plot of the learned time offsets over training iterations on \texttt{coffee\_martini}. This scene has an unsynchronized view as shown in \fref{fig:teaser}a.
The time offsets are initialized as zero and converge to the visually correct offset. As the common scenes do not provide ground truth offsets, we report errors on synthetic dynamic scenes with the ground truth offsets in the following experiments. We note that the optimization of time offsets does not require additional loss functions or regularizers.

When we assess the reconstruction of the scene captured from a test view, even the test view can be unsynchronized. For rendering a video from a novel view, we can customize the time offset $\offset_\text{test}$. Our method allows to optimize $\offset_\text{test}$ with the frozen trained model such that the potential time offset of the test view can be resolved. To fully exploit the advantage of our method, we optimize the time offset for the test view when we report the test view performance.

\begin{figure*}[tb!]
  \centering
  \includegraphics[width=0.9\linewidth]{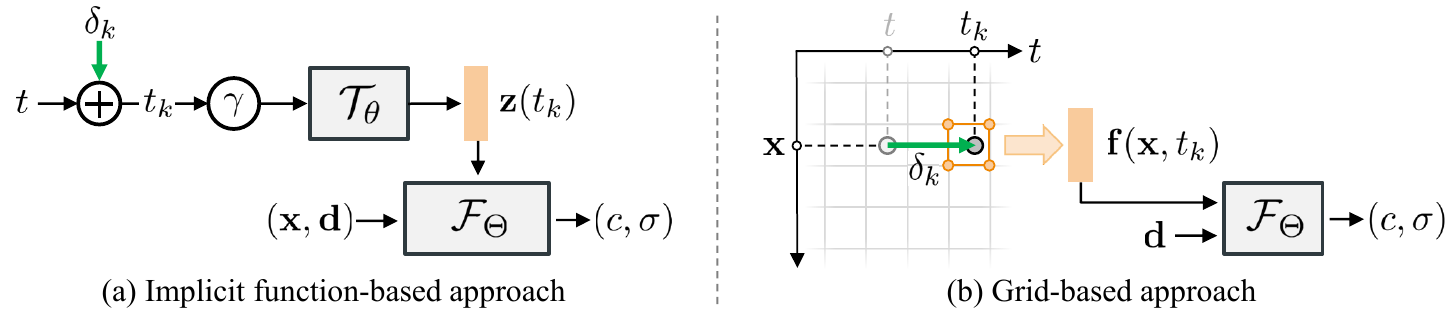}
  \caption{\textbf{Continuous temporal embedding.}  (a) We present an implicit function-based approach for the methods utilizing per-frame temporal embeddings. We add time offset $\offset_k$ of camera $k$ to time input $t$. $\mathcal{T}_\theta$ is the implicit function for mapping calibrated time into temporal embedding $\mathbf{z}$. (b) We query the embedding at the calibrated time $t_k$ on grid-based models. 
  Bilinear interpolation naturally allows continuous temporal embedding.}
\label{fig:method}
\end{figure*}


\subsection{Implicit Function for Temporal Embedding}
\label{sec:implicit}
Various dynamic NeRFs \cite{park2021hypernerf,li2022neural,wang2022mixed} explicitly learn latent embeddings of individual frames to represent the temporal variations of dynamic scenes. These latent embeddings do not represent the moments between frames nor are their interpolation is not guaranteed to produce smooth dynamics. 
Furthermore, the number of embeddings also increases as the video becomes longer, requiring more memory usage.

Instead of optimizing hundreds of individual embedding vectors for all frames, we train an implicit neural representation $\mathcal{T}_{\theta}$ that produces the temporal embedding $\mathbf{z}(t)$ for an arbitrary time input $t$ (\fref{fig:method}a). First, we encode a normalized time input $t$ using a set of sinusoidal functions:
\begin{equation}
  \pe(t,L) = \left[\sin(2^0\pi t),\cdots,\sin(2^{L-1}\pi t), \cos(2^{L-1}\pi t) \right]^{\mathrm T}.
  \label{eq:timepe}
\end{equation}

Similar to previous observations \cite{mildenhall2021nerf,tancik2020fourier}, where inputs encoded with high-frequency functions lead to a better fit for high-frequency variations in data, this mapping assists $\mathcal{T}_{\theta}$ in capturing the movement of the scene. Thus, \eref{eq:func_ours} is modified as follows: 
\begin{equation}
  \mathcal{F}_\mlpparam : (\pos, \dir, \mathbf{z}(t_k)) \rightarrow (\radiance, \sigma), \; \text{where }\; \mathbf{z}(t) = \mathcal{T}_{\theta}(\pe(t,L)),
  \label{eq:timemlp}
\end{equation}
where $t_k$ is the time input shifted by the per-camera time offset used in \eref{eq:func_ours}. To capture the rapid scene motion, we set $L=10$ in all experiments. We select MixVoxels as a baseline for using per-frame temporal latents and compare it with our \mbox{Sync-MixVoxels} in \Sref{sec:results}.


\subsection{Grid-based Models with Time Offsets}
\label{sec:grid}
Grid-based dynamic NeRFs \cite{fridovich2023k,park2023temporal,attal2023hyperreel} calculate latent vectors from the feature grid to feed the latents to the time-dependent function $\mathcal{F}_\mlpparam$. Specifically, for a given 4D coordinates $(\pos, t)$, the latent representation $\mathbf{f}(\pos,t)$ is obtained by linearly interpolating feature vectors assigned to each vertex of the grid to which the coordinates belong.

We use camera-specific time $t_k$ instead of the original time $t$. In grid-based models, \eref{eq:func_ours} is modified as follows:

\begin{equation}
  \mathcal{F}_\mlpparam : (\mathbf{f}(\pos, t_k), \dir) \rightarrow (\radiance, \sigma), \; \text{where }\; \mathbf{f}(\pos, t) = \text{Grid}(\pos, t),
  \label{eq:grid}
\end{equation}

and $\text{Grid}(\cdot)$ denotes the interpolation of grid vectors surrounding given coordinates. 
{\fref{fig:method}b illustrates how our method modifies the sampling in the grid by $\offset_k$. We select K-Planes as a baseline for using grid-based representation and compare it with our \mbox{Sync-K-Planes}.

        
\section{Experiments}
\label{sec:results}

In this section, we validate the effectiveness of our method using MixVoxels and K-Planes as baselines.  \sref{sec:exp:setting} describes the datasets and evaluation metrics. \sref{sec:exp:quality} shows that our method significantly improves the baselines regarding the shape of moving objects and overall reconstruction. \sref{sec:exp:offset} validates the accuracy of the found time offsets. \sref{sec:various} demonstrates the robustness of our method against different levels of unsynchronization including synchronized settings.

\begin{figure}[t!]
  \centering
  \includegraphics[width=0.9\linewidth]{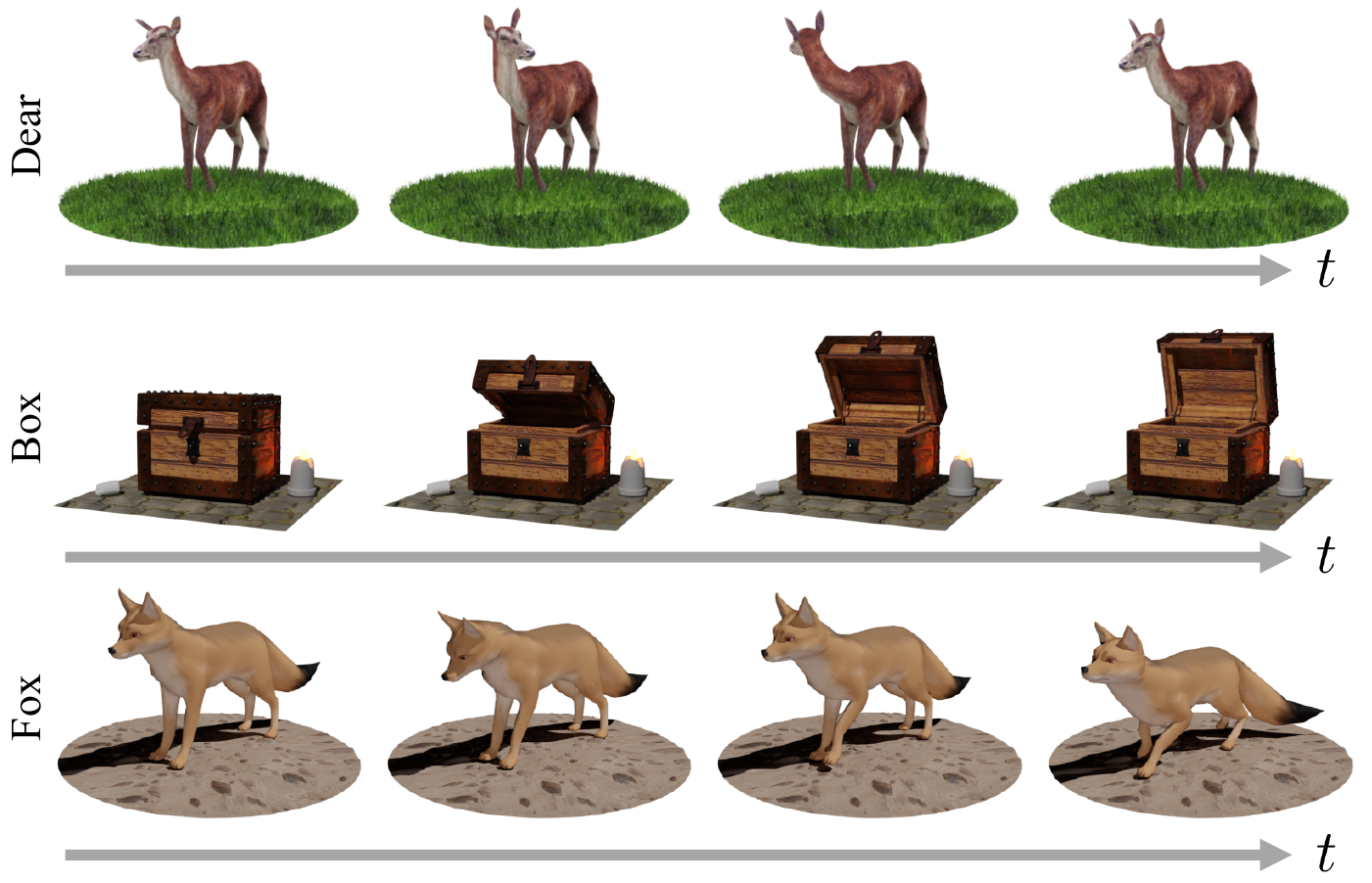}
  \caption{\textbf{Snapshots of the Unsynchronized Dynamic Blender Dataset.} We show exemplar frames of a video from our multi-view synthetic dataset.
  }
\label{fig:blender}
\end{figure}

\begin{figure*}[t!]
  \centering
  \includegraphics[width=0.95 \linewidth]{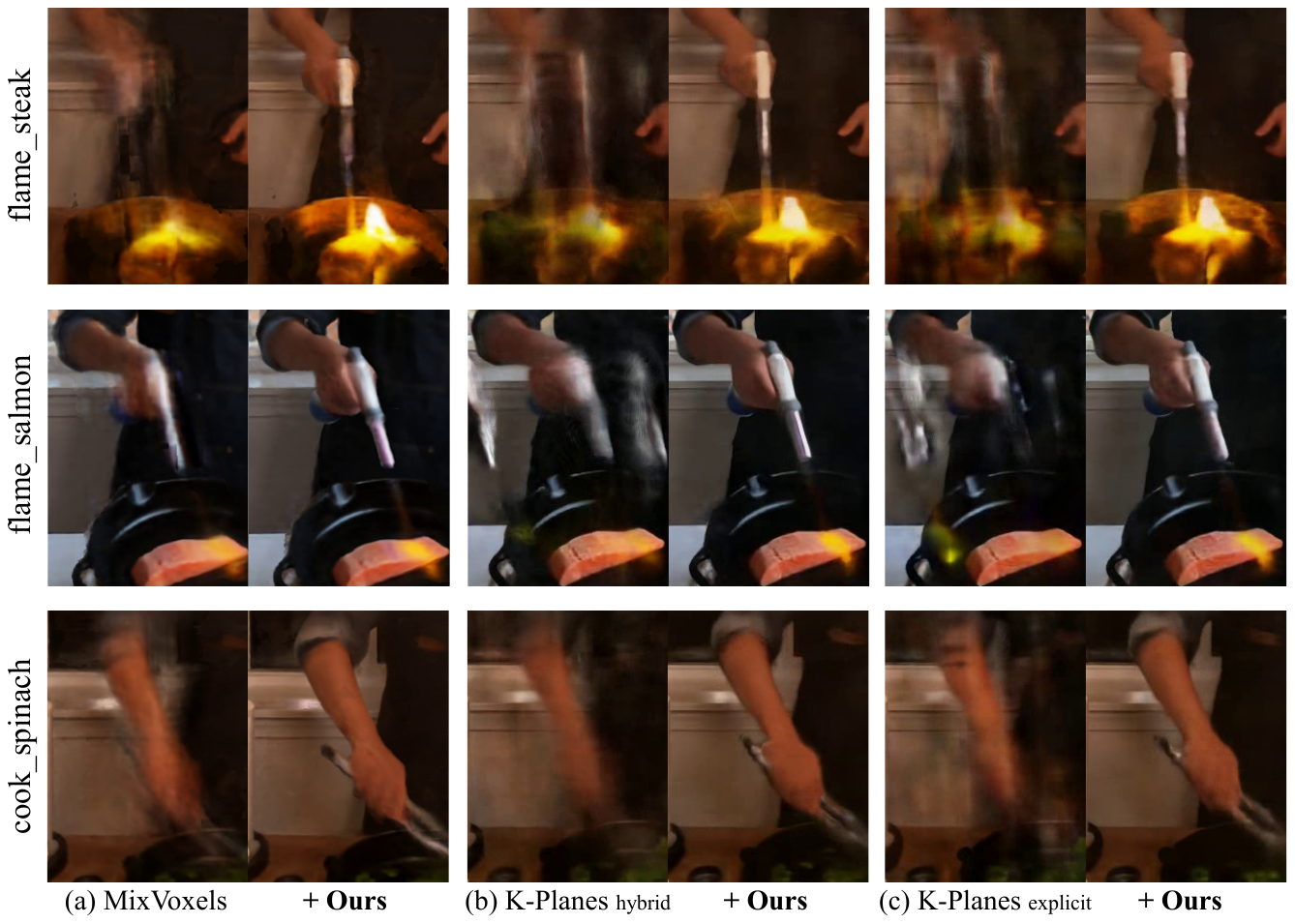}
  \caption{\textbf{Cropped renderings on Unsynchronized Plenoptic Video Dataset.} While the baselines produce severe artifacts, employing our method on them resolves the problem.}
\label{fig:real_main}
\end{figure*}


\subsection{Evaluation Settings}
\label{sec:exp:setting}
\subsubsection{Unsynchronized Datasets.}
The Plenoptic Video Dataset \cite{li2022neural} contains six challenging real-world scenes with varying degrees of dynamics. Its multi-view videos are roughly synchronized except \texttt{coffee\_martini} scene. In order to simulate an in-the-wild unsynchronized capturing environment, we modify this dataset to be unsynchronized by randomly translating along the temporal axis. The time offsets are sampled from a normal distribution with zero mean and a standard deviation of 5, and then rounded to be integers. More details are in Appendix A.

Although the Plenoptic Video Dataset is synchronized, we cannot prepare ground truth offsets because the synchronization is not perfect. As a solution, we create an Unsynchronized Dynamic Blender Dataset as the following process. We start from free public Blender assets with motion, namely \texttt{box}, \texttt{fox}, and \texttt{deer}. These assets are rendered on similar camera setups. Subsequently, we translate the rendered videos according to random time offsets drawn from the aforementioned normal distribution. Then we have access to the ground truth time offsets because these rendered videos are perfectly synchronized. All videos are 10 seconds long and captured in 14 fixed frontal-facing multi-view cameras at a frame rate of 30 FPS following the Plenoptic Video Dataset. Example frames are shown in \fref{fig:blender}. The dataset is publicly available.


\subsubsection{Baselines.}
We employ MixVoxels and K-Planes as baselines and adopt our method upon them, namely, Sync-MixVoxels and Sync-K-Planes. We select the baselines because they are the latest among the methods with per-frame temporal latent and grid-based temporal representation.


\subsubsection{Evaluation Metrics.}
We evaluate the rendering quality using the following quantitative metrics. To quantify the pixel color error, we report PSNR (peak signal-to-noise ratio) between rendered video and the ground truth. To consider perceived similarity, we report SSIM \cite{wang2004image}. To measure higher-level perceptual similarity, we report LPIPS \cite{zhang2018unreasonable} using VGG and AlexNet. Higher values for PSNR and SSIM, and lower values for LPIPS indicate better visual quality. To measure the accuracy of the time offsets, the mean absolute error (MAE) between the found offsets and the ground truth offsets is measured in seconds.

\begin{figure*}[htb!]
  \centering
  \includegraphics[width=0.9 \linewidth]{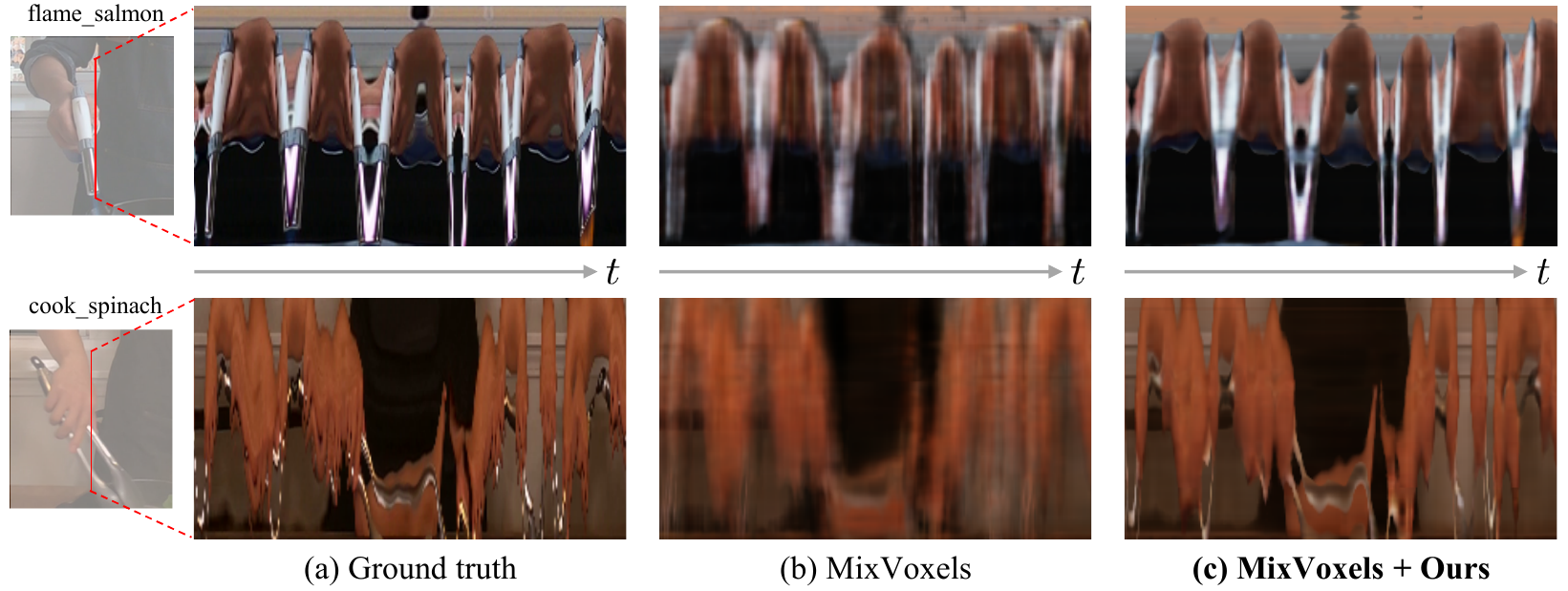}
  \caption{\textbf{Spatio-temporal Images.} Each column in the image represents the rendered pixels on the fixed vertical line at a moment. The fixed vertical line with 150 pixels is shown in the leftmost image patch. We render all frames in the video, producing 270-pixel-wide spatiotemporal images.
  Our results are much clearer than the baseline.}
\label{fig:plane}
\end{figure*}


\subsection{Rendering quality}
\label{sec:exp:quality}
We evaluate the effectiveness of our method over the baselines by observing rendering quality. We highly recommend watching the result video on our project page.

\subsubsection{Unsynchronized Plenoptic Video Dataset.}
\fref{fig:real_main} compares ours with the baselines on the Unsynchronized Plenoptic Video Dataset. All the baselines produce severe artifacts in dynamic parts and perform even worse on scenes with larger motion. Opposed to the reported performance of K-Planes outperforming MixVoxels on synchronized dataset, K-Planes suffers more on dynamic parts in unsynchronized setting: the hand and the torch are rendered in multiple places in K-Planes. Refer to per-scene results in Appendix F.1.

However, our method successfully corrects artifacts in both baselines. Additionally, in \fref{fig:plane}, we visually demonstrate the comparison of spatio-temporal images on dynamic regions. Each column in an image represents the rendered rays on the fixed vertical line at a time. Horizontally concatenating columns of all frames from the test view constructs a spatio-temporal image as a whole. Our method exhibits significantly clearer spatio-temporal images compared to the baseline (\fref{fig:plane}b-c).

\begin{table}[t]
\label{}
\resizebox{\columnwidth}{!}{
\setlength\tabcolsep{4pt}

\begin{tabular}{rc|ccccc}

\multicolumn{2}{c|}{\diagbox[width=\widthof{Sync-K-Planes}+\widthof{\scriptsize{explicit}}+4\tabcolsep+\arrayrulewidth\relax, height=1.5\line]{model}{metric}} & PSNR         & SSIM          & \makecell{LPIPS{\scriptsize{alex}}}  & \makecell{LPIPS{\scriptsize{vgg}}}   
\\ \hline

MixVoxels  &       & 29.96                    & 0.9059          & 0.1669          & 0.2648          \\
Sync-MixVoxels && \textbf{30.53}  & \textbf{0.9101} & \textbf{0.1570} & \textbf{0.2575} \\ \hline

K-Planes &\scriptsize{hybrid} &        29.16 & 0.9120 & 0.1278 & 0.2222 \\
Sync-K-Planes &\scriptsize{hybrid} & \textbf{30.44}                        & \textbf{0.9243}          & \textbf{0.1064}          & \textbf{0.1989}          \\ \hline

K-Planes  &\scriptsize{explicit}         & 28.51                        & 0.9042          & 0.1484          & 0.2438          \\
Sync-K-Planes & \scriptsize{explicit} & \textbf{29.97}                        & \textbf{0.9223}          & \textbf{0.1144}          & \textbf{0.2103}         
\end{tabular}
}
\caption{\textbf{Average performance in the \emph{test view} on Unsynchronized Plenoptic Video Dataset.} Our method improves all the baselines, even achieving performance similar to synchronized setting in \tref{tab:sync_table}.}
 \label{tab:real_test}
\end{table}

\tref{tab:real_test} reports quantitative metrics on the \emph{test} views of the unsynchronized Plenoptic Video Dataset. We report the average across all scenes and refer per-scene performance to the Appendix F.1.
Our method improves the baselines in all cases nearly to the performance on synchronized setting (\sref{sec:exp:sync}). The above values are the result after optimizing the time offset in the test view. The previous results are reported in Appendix B.

\tref{tab:real_train} reports the same metrics on the \emph{training} views. The baselines struggle to reconstruct even the training views. On the other hand, adopting our method on the baselines fixes the problem.

\begin{table}[t]

\resizebox{\columnwidth}{!}{
\setlength\tabcolsep{4pt}
\begin{tabular}{rc|ccccc}

\multicolumn{2}{c|}{\diagbox[width=\widthof{Sync-K-Planes}+\widthof{\scriptsize{explicit}}+4\tabcolsep+\arrayrulewidth\relax, height=1.5\line]{model}{metric}} & PSNR         & SSIM          & \makecell{LPIPS{\scriptsize{alex}}}  & \makecell{LPIPS{\scriptsize{vgg}}}   
\\ \hline

MixVoxels &&        31.13 & 0.9115 & 0.1565 & 0.2565 \\
Sync-MixVoxels && \textbf{31.87} & \textbf{0.9162} & \textbf{0.1457} & \textbf{0.2495} \\ \hline

K-Planes &\scriptsize{hybrid} & 29.25 & 0.9013 & 0.1548 & 0.2477 \\
Sync-K-Planes &\scriptsize{hybrid} & \textbf{30.59} & \textbf{0.9221} & \textbf{0.1118} & \textbf{0.2042} \\ \hline

K-Planes &\scriptsize{explicit} &        29.64 & 0.9095 & 0.1461 & 0.2386 \\
Sync-K-Planes &\scriptsize{explicit} & \textbf{30.79} & \textbf{0.9238} & \textbf{0.1147} & \textbf{0.2050} \\         
\end{tabular}
}
\caption{\textbf{Average performance over all \emph{training views} on Unsynchronized Plenoptic Video Dataset.} Our method improves all the baselines.}  
\label{tab:real_train}
\end{table}

\begin{figure*}[htb!]
  \centering
  \includegraphics[width=1\linewidth]{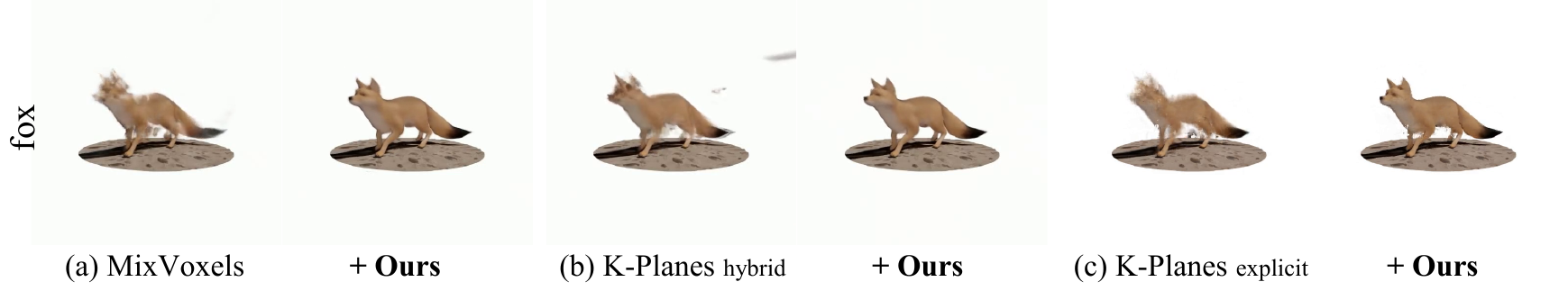}
  \caption{\textbf{Qualitative results on Unsynchronized Dynamic Blender Dataset} Visual comparison of MixVoxels, K-Planes hybrid, K-Planes explicit and our method on them with \texttt{fox} scene.}
  \label{fig:blender_qual}
\end{figure*}


\subsubsection{Unsynchronized Dynamic Blender Dataset.}
\fref{fig:blender_qual} compares the baselines and our approaches on $\texttt{fox}$ scene from Unsynchronized Dynamic Blender Dataset. While MixVoxels and K-Planes struggle on the motion and object boundaries, our Sync-MixVoxels and Sync-K-Planes shows the clearer face of the fox. 

\tref{tab:synthetic_test} reports quantitative metrics on the \emph{test} views of the Unsynchronized Dynamic Blender Dataset. We report the average across all scenes and refer to per-scene performance in Appendix F.2.
Our method improves the baselines in all cases. 


\subsection{Time Offset Accuracy}
\label{sec:exp:offset}
We demonstrate that the time offsets found by our method are highly accurate. \tref{tab:synthetic_error} reports the mean absolute error (MAE) between the optimized time offsets and the ground truth offsets on the Unsynchronized Dynamic Blender Dataset. The average error is approximately 0.01 seconds, which corresponds to less than one-third of a frame.

\begin{table}[t]

\resizebox{\columnwidth}{!}{
\setlength\tabcolsep{4pt}

\begin{tabular}{rc|ccccc}

 \multicolumn{2}{c|}{\diagbox[width=\widthof{Sync-K-Planes}+\widthof{\scriptsize{explicit}}+4\tabcolsep+\arrayrulewidth\relax, height=1.5\line]{model}{metric}} & PSNR         & SSIM          & \makecell{LPIPS{\scriptsize{alex}}}  & \makecell{LPIPS{\scriptsize{vgg}}}   
\\ \hline

MixVoxels  &       & 31.06	& 0.9753	& 0.0237	& 0.0339 \\
Sync-MixVoxels & & \textbf{37.11}	& \textbf{0.9841}	& \textbf{0.0120}	& \textbf{0.0226} \\ \hline

K-Planes  &\scriptsize{hybrid} & 32.66	& 0.9771	& 0.0175	& 0.0244 \\
Sync-K-Planes &\scriptsize{hybrid} & \textbf{39.40}	& \textbf{0.9863}	& \textbf{0.0066}	& \textbf{0.0131} \\ \hline

K-Planes  &\scriptsize{explicit} & 32.15	& 0.9751	& 0.0246	& 0.0335\\
Sync-K-Planes & \scriptsize{explicit} & \textbf{39.35}	& \textbf{0.9865}	& \textbf{0.0072}	& \textbf{0.0140}         
\end{tabular}
}
\caption{\textbf{Average performance in the \emph{test view} on Unsynchronized Dynamic Blender Dataset.} Our method improves all the baselines.} 
\label{tab:synthetic_test}
\end{table}

\begin{table}[!ht]
\centering
\setlength\tabcolsep{4pt}
\begin{tabular}{rc|c}

 && MAE \scriptsize{(seconds)}
\\ \hline

Sync-MixVoxels && 0.0154 \\ \hline

Sync-K-Planes &\scriptsize{hybrid} & 0.0156 \\ \hline

Sync-K-Planes & \scriptsize{explicit} & 0.0229
\end{tabular}
\caption{\textbf{MAE between predicted offset and ground truth on Unsynchronized Dynamic Blender Dataset.} Our method accurately finds time offsets achieving MAE of approximately 0.01 seconds.}
\label{tab:synthetic_error}
\end{table}


\subsection{Versatility to Various Scenarios} 
\label{sec:various}
\subsubsection{Various Unsynchronization Lengths.}
We evaluate the performance of all methods under different lengths of unsynchronization, namely $1.5\times$ and $2\times$ long offsets. \tref{tab:robust_table} shows that our method consistently improves the performance of the baselines. We note that the unsynchronization deteriorates K-Planes more than MixVoxels even though K-Planes outperform MixVoxels on synchronized datasets. Nevertheless, our method successfully reflects their ranking in synchronized setting to unsynchronized setting.


\subsubsection{Synchronization Setting.}
\label{sec:exp:sync}

We verify that our method improves the baselines even on synchronized settings. Although the gaps between the baselines and ours are smaller than unsynchronized setting, this improvement implies that the dataset assumed to be synchronized is not perfectly synchronized and even tiny offsets less than a frame are correctly found by our method.

\begin{table}[!ht]

\resizebox{\columnwidth}{!}{
\setlength\tabcolsep{4pt}
\begin{tabular}{rc|ccc|ccc}

&& \multicolumn{3}{c|}{1.5$\times$} & \multicolumn{3}{c}{2.0$\times$}                    \\ \hline
\multicolumn{2}{c|}{\diagbox[width=\widthof{Sync-K-Planes}+\widthof{\scriptsize{explicit}}+4\tabcolsep+\arrayrulewidth\relax]{model}{scene}} & \makecell{cook\\spinach} & \makecell{flame\\salmon} & fox & \makecell{cook\\spinach} & \makecell{flame\\salmon} & fox \\ \hline

MixVoxels && 30.21 & 27.27 & 31.10& 30.48 & 27.66 & 29.81\\
Sync-MixVoxels && \textbf{31.44} & \textbf{28.59} & \textbf{36.15}& \textbf{31.50} & \textbf{28.74} & \textbf{35.58}\\ \hline
K-Planes &\scriptsize{hybrid} & 29.93&26.11&31.85&29.13&26.01&29.71\\
Sync-K-Planes &\scriptsize{hybrid} & \textbf{31.71} & \textbf{28.67} & \textbf{40.31} & \textbf{31.85} & \textbf{28.22} & \textbf{40.36}\\ \hline
K-Planes &\scriptsize{explicit} & 28.93&24.71&31.46&28.46&24.83&29.10\\
Sync-K-Planes &\scriptsize{explicit} & \textbf{31.12} & \textbf{28.41} & \textbf{40.47} & \textbf{31.16} & \textbf{27.09} & \textbf{40.29} \\

\end{tabular}
\textbf{}}
\caption{\textbf{Average PSNR in the \emph{test view} with varaious unsynchonization lengths.} Our method is robust to different lengths of unsynchronization. }
\label{tab:robust_table}
\end{table}
\begin{table}[!ht]

\resizebox{\columnwidth}{!}{
\setlength\tabcolsep{4pt}
\begin{tabular}{rc|ccccc}

\multicolumn{2}{c|}{\diagbox[width=\widthof{Sync-K-Planes}+\widthof{\scriptsize{explicit}}+4\tabcolsep+\arrayrulewidth\relax, height=1.5\line]{model}{metric}} & PSNR         & SSIM          & \makecell{LPIPS{\scriptsize{alex}}}  & \makecell{LPIPS{\scriptsize{vgg}}}   
\\ \hline

MixVoxels &&        30.39 & 0.9100 & 0.1577 & 0.2586 \\
Sync-MixVoxels && \textbf{30.41} & \textbf{0.9104} & \textbf{0.1559} & \textbf{0.2564} \\ \hline

K-Planes &\scriptsize{hybrid} & 30.40 & \textbf{0.9257} & \textbf{0.1044} & \textbf{0.1980} \\
Sync-K-Planes &\scriptsize{hybrid} & \textbf{30.56} & 0.9246 & 0.1059 & 0.1998 \\ \hline

K-Planes &\scriptsize{explicit} &        30.04&0.9229&0.1131&0.2083 \\
Sync-K-Planes &\scriptsize{explicit} & \textbf{30.14} & \textbf{0.9237} & \textbf{0.1121} & \textbf{0.2072} \\         
\end{tabular}
}
\caption{\textbf{Average performance in the \emph{test view} on Synchronized Plenoptic Video Dataset.} Our method enhances performance on synchronized setting by resolving the small temporal gaps.}
\label{tab:sync_table}
\end{table}

\section{Conclusion}
Our work is the first attempt to train dynamic NeRFs on unsynchronized multi-view videos. We have shown that the existing dynamic NeRFs deteriorate when the videos are not synchronized. As its reason lies in the previous method using a single temporal latent embedding for a multi-view frame, we introduce time offsets for individual views such that the videos can be synchronized by the offsets. We jointly optimize the offsets along with NeRF, using a typical reconstruction loss. Our method, Sync-NeRF, is versatile to various types of existing dynamic NeRFs, including Sync-MixVoxels, Sync-K-Planes, and more. It consistently improves the reconstruction on both unsynchronized and synchronized settings.

\paragraph{Discussion} Although our method drives dynamic NeRFs closer to in-the-wild captures, we still lack the means to generalize to the causal handheld cameras. We suggest it as an interesting future research topic. While our method does not introduce additional computational complexity to grid-based temporal embedding, implicit function-based temporal embeddings require an additional training time and memory for training the function. Nevertheless, in the inference phase, the temporal embeddings can be pre-computed for all frames leading to negligible overhead.

\section*{Acknowledgments}
This work is supported by the Institute for Information \& Communications Technology Planning \& Evaluation (IITP) grant funded by the Korea government(MSIT)
(No. 2017-0-00072, Development of Audio/Video Coding and Light Field Media Fundamental Technologies for Ultra Realistic Tera-media)

\bibliography{aaai24}

\clearpage

\title{Supplemental Material for Sync-NeRF}

\renewcommand{\thetable}{S\arabic{table}}
\renewcommand{\thefigure}{S\arabic{figure}}
\setcounter{figure}{0}
\setcounter{table}{0}



\begin{appendix}
\section{Additional Details}
\label{sup:details}


\subsection{Training Details}

\subsubsection{Sync-MixVoxels} The implicit function $\mathcal{T}_{\theta}$ is a 2-layer multi-layer perceptron (MLP) with 512 hidden units. Due to memory resource constraints, we use 64 and 128 ray batch sizes that query all frames of Unsynchornized Dynamic Blender Dataset and Unsynchronized Plenoptic Video Dataset, respectively. We train both MixVoxels and Sync-MixVoxels for 60K iterations. The time offset $\delta_{k}$ for Sync-MixVoxels starts training after the first 10K iterations for initial training stability. 
We set the learning rate of implicit function and time offset to be $5\times$ and $0.5\times$ of the MixVoxels backbone, respectively.
The rest of the details follow the MixVoxels paper. 

\subsubsection{Sync-K-Planes}
We train K-Planes with a batch size of 4096 rays. Two versions of K-Planes, hybrid and explicit, are trained for 90K and 120K iterations, respectively. K-Planes applies ray sampling based on the temporal difference (IST) weight map proposed by DyNeRF. We apply the ISG sampling strategy for the first 50K iterations with time offset and then apply the IST strategy for both K-Planes and Sync-K-Planes in Plenoptic Video Dataset. We set the time offset learning rate to be $0.1\times$ of the K-Planes. We roughly change the normalization range of time $t$ to $[-0.8, 0.8]$ so that the camera-specific time $t_k=t+\delta_k$ does not deviate from the feature planes. For a comparison of quantitative results for the temporal axis resolution of the K-Planes, see \sref{sup:time_norm}.

\subsubsection{Computational complexity}
\tref{tab:supp_comp} reports the computational complexity and the number of model parameters on \texttt{flame\_salmon} scene. Our method brings negligible or tolerable differences in rendering and training time, and the number of model parameters.

\begin{table}[!ht]
\centering
\resizebox{0.95\columnwidth}{!}{
\setlength\tabcolsep{4pt}

\begin{tabular}{rc|cccccc}

& & Rendering & Training & \# Params \\ \hline

MixVoxels & & 2.3 min & 1.9 hrs & 130.7 M \\
Sync-MixVoxels & & 2.5 min & 2.5 hrs & 131.8 M \\ \hline

K-Planes &\scriptsize{hybrid} & 15.0 min & 1.9 hrs  & 27.0 M\\
Sync-K-Planes &\scriptsize{hybrid} & 15.2 min & 2.0 hrs & 27.0 M\\ \hline 

K-Planes &\scriptsize{explicit} & 21.7 min & 3.7 hrs & 50.7 M \\
Sync-K-Planes & \scriptsize{explicit} & 22.0 min & 3.8 hrs & 50.7 M \\ \hline 
\end{tabular}
}
\caption{\textbf{Comparison of computational complexity and model size on} \texttt{flame\_salmon} \textbf{scene.} Adopting our method on baselines leads to negligible additional computational cost.
}
\label{tab:supp_comp}
\end{table}

\subsubsection{Misc.}
All experiments are trained on a single NVIDIA A6000. Since the target datasets are forward-facing scenes, we use normalized device coordinates in Plenoptic Video Dataset.


\subsection{Dataset Details}
\label{sup:dataset}

\subsubsection{Plenoptic Video Dataset}
The Plenoptic Video Dataset shows imperfect synchronization even though it is synchronized using a hardware device. As shown in the \fref{fig:supp_sync}, fast dynamic contents such as flames show different visual appearances in the same frame number of training videos. Nevertheless, our method accurately estimates time offsets and reconstructs the dynamic scene without accurate synchronization.

\begin{figure}[h!]
  \centering
  \includegraphics[width=1 \linewidth]{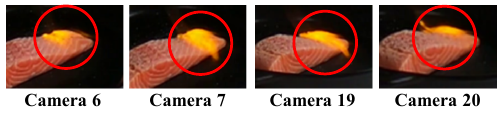}
  \caption{\textbf{Imperfect hardware synchronization.} The flame of the \texttt{frame\_salmon} scene shows different appearances at the same frame number.}
\label{fig:supp_sync}
\end{figure}

\subsubsection{Dynamic Blender Dataset}
To create a dynamic Blender dataset, we use free 3D models that provide animation
\footnote{\texttt{dear}: https://www.turbosquid.com/3d-models/3d-deer-animation-model/1012108 \\ \texttt{box}: https://www.turbosquid.com/3d-models/treasure-box-candles-model-2010834 \\ \texttt{fox}: https://www.turbosquid.com/3d-models/animation-3d-model-1589869}.
We add a floor under the object for each scene and slightly modify the animation to render it 300 frames at $512 \times 512$ resolution. We do not apply downsampling for training. 13 cameras are uniformly sampled on a spherical surface to render a forward-facing scene, and one camera is placed in the center of the surface to be used as a test view. All cameras share intrinsic parameters.

\subsubsection{Misc.}
To create unsynchronized datasets, we translate each video along the temporal axis and then use the first 270 frames of overlapping intervals as training data. In \sref{sec:various}, we use the first 255 and 240 frames for $1.5\times$ and $2.0\times$ settings, respectively.


\section{Test View Offset Optimization}
\label{sup:testoptim}
To fully exploit the advantages of camera-specific time offset, we optimize the test view time offset $\delta_{\text{test}}$ to evaluate our models. We optimize the time offset for 200 and 1K iterations on MixVoxels and K-Planes, respectively, and then perform the evaluation. We train only the time offset while freezing the entire network. \tref{tab:supp_testoptim} shows that optimizing the time offset in the test view improves performance slightly.
\begin{table}[!ht]
\resizebox{\columnwidth}{!}{
\setlength\tabcolsep{4pt}

\begin{tabular}{rl|cccc}
\multicolumn{2}{r|}{\textbf{average}} & PSNR & SSIM & \makecell{LPIPS{\scriptsize{alex}}} & \makecell{LPIPS{\scriptsize{vgg}}}

\\ \hline
\multicolumn{2}{l|}{Sync-MixVoxels} & \textbf{30.53} & \textbf{0.9101} & \textbf{0.1570} & \textbf{0.2575} \\

\multicolumn{2}{l|}{$\quad \llcorner \text{w/o} \; \delta_{\text{test}}$} & 30.29 & 0.9091 & 0.1574 & 0.2579          \\ \hline

Sync-K-Planes & \scriptsize{hybrid} & \textbf{30.44} & \textbf{0.9243} & \textbf{0.1064} & \textbf{0.1989} \\
\multicolumn{2}{l|}{$\quad \llcorner \text{w/o} \; \delta_{\text{test}}$} & 30.25 & 0.9235 & 0.1067 & 0.1992          \\ \hline

Sync-K-Planes & \scriptsize{explicit} & \textbf{29.97} & \textbf{0.9223} & \textbf{0.1144} & \textbf{0.2103} \\
\multicolumn{2}{l|}{$\quad \llcorner \text{w/o} \; \delta_{\text{test}}$} & 29.81 & 0.9214 & 0.1146 & 0.2104     

\end{tabular}
}
\caption{\textbf{Comparison with test view offset optimization on Unsynchronized Plenoptic Video Dataset.} Test view offset optimization enables more accurate evaluation and improves our quantitative results.
}
\label{tab:supp_testoptim}
\end{table}


\section{Spatio-Temporal Image of K-Planes}
\label{sup:kple_img}

\begin{figure*}[htb!]
  \centering
  \includegraphics[width=0.85\linewidth]{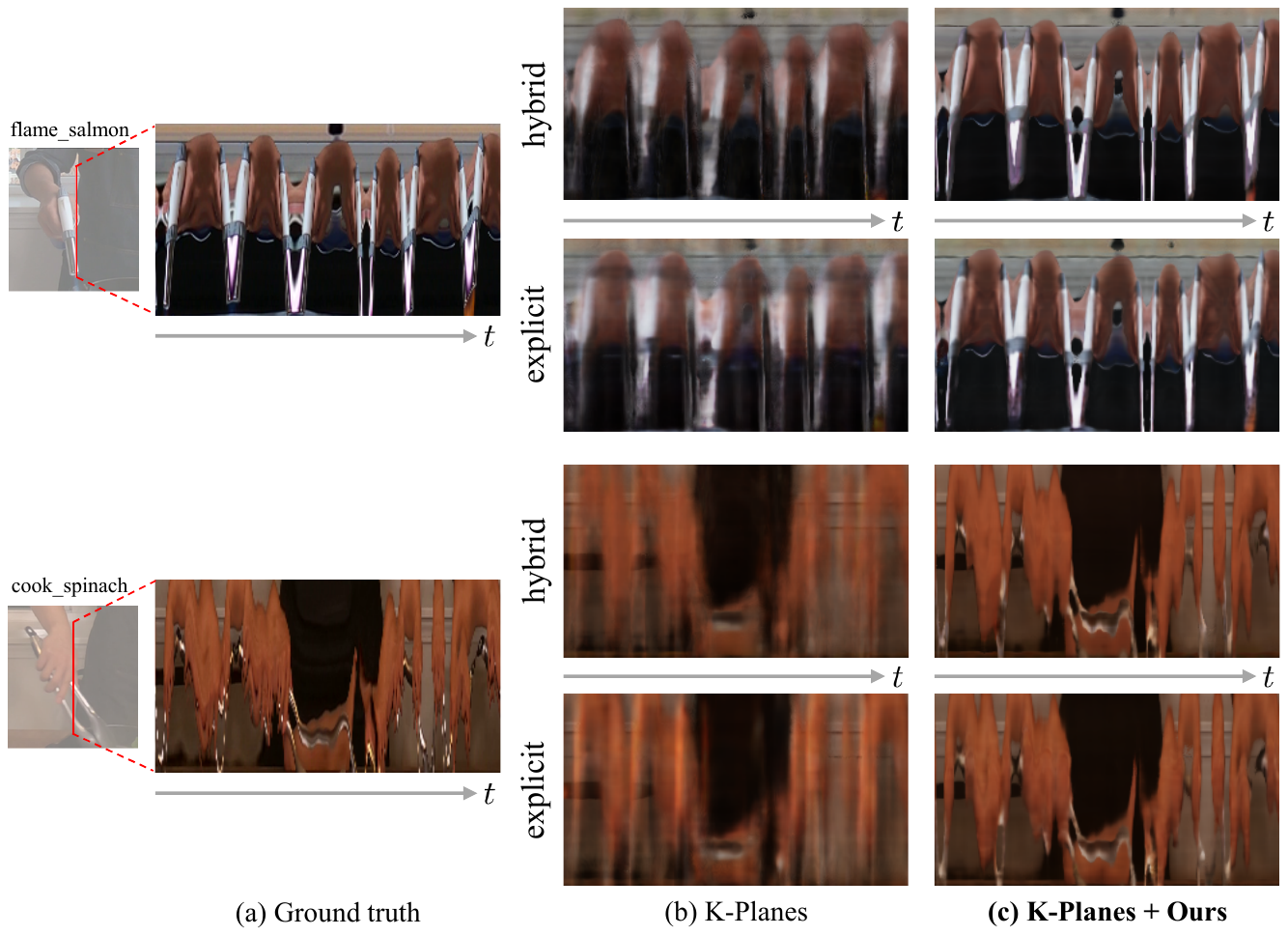}
  \caption{\textbf{Spatiotemporal Images of K-Planes.} Our results are much clearer than the baseline. }
  \label{fig:supp_spatialtemporal}
\end{figure*}

In \fref{fig:supp_spatialtemporal}, we show the spatiotemporal images for K-Planes and Sync-K-Planes on \texttt{flame\_salmon} scene and \texttt{cook\_spinach} scene. Applying our method, both hybrid and explicit versions of K-Planes render much clearer results.


\section{Sinusoidal Functions for Time}
\label{sec:pe_abl}

For our implicit function-based approach, we encode normalized time input using a set of sinusoidal functions to better represent the movement of the scene. 
\tref{tab:supp_pe} show the results of Sync-MixVoxels on \texttt{flame\_salmon} scene for different $L$s. With $L\!\!=\!\!0$, the model fails to represent dynamic scenes, and $L\!\!=\!\!5$ has a lower quality of dynamic regions than $L\!\!=\!\!10$, so we set $L\!\!=\!\!10$ as the default.

\begin{table}[!th]
\begin{tabular}{l|cccc}
$L$ for time &  PSNR & SSIM & \makecell{LPIPS{\scriptsize{alex}}} & \makecell{LPIPS{\scriptsize{vgg}}}
 
\\ \hline
0 & 22.26 & 0.8411 & 0.2336 & 0.3001 \\
5  & 28.80 & 0.8784 & \textbf{0.2006} & 0.2741 \\
10 (default) & \textbf{28.85} & \textbf{0.8793} & 0.2022 & \textbf{0.2740}

\end{tabular}

\caption{\textbf{Comparision of different $L$s for sinusoidal functions on} \texttt{flame\_salmon} \textbf{scene.} Using a high $L$ value provides better dynamic region quality, so we set $L=10$ as the default value.}
\label{tab:supp_pe}
\end{table}


\section{Temporal Axis Resolution of K-Planes}
\label{sup:time_norm}
    
In our K-Planes experiments, the resolution of the temporal axis is lowered 
because the normalization range of time is reduced from $[-1.0, 1.0]$ to $[-0.8, 0.8]$.
In \tref{tab:supp_resolution}, we report the results of increasing the temporal-axis resolution of the feature grid up to its original setting. As shown in the table, reducing the normalization range does not harm the overall performance.

\begin{table}[!ht]
\centering\resizebox{\columnwidth}{!}{
\begin{tabular}{rl|cccc}
 &  & PSNR & SSIM & \makecell{LPIPS{\scriptsize{alex}}} & \makecell{LPIPS{\scriptsize{vgg}}}
 
\\ \hline
\multicolumn{2}{r|}{K-Planes \scriptsize{hybrid}} & 26.19 & 0.8757 & 0.1698 & \textbf{0.2559} \\

\multicolumn{2}{r|}{$\llcorner$ $t$-axis res $\uparrow$} & \textbf{26.56} & \textbf{0.8760} & \textbf{0.1694} & 0.2589 \\ \hline

\multicolumn{2}{r|}{K-Planes \scriptsize{explicit}} & \textbf{25.61} & \textbf{0.8675} & \textbf{0.1891} & \textbf{0.2728} \\

\multicolumn{2}{r|}{$\llcorner$ $t$-axis res $\uparrow$} & 25.31 & 0.8631 & 0.1963 & 0.2821 \\ \hline

\multicolumn{2}{r|}{Sync-K-Planes \scriptsize{hybrid}} & \textbf{29.17} & \textbf{0.9046} & \textbf{0.1259} & \textbf{0.2112} \\
\multicolumn{2}{r|}{$\llcorner$ $t$-axis res $\uparrow$} & 28.95 & 0.9027 & 0.1290 & 0.2131 \\ \hline

\multicolumn{2}{r|}{Sync-K-Planes \scriptsize{explicit}} &  28.56 & \textbf{0.9048} & \textbf{0.1295} & \textbf{0.2114} \\
\multicolumn{2}{r|}{$\llcorner$ $t$-axis res $\uparrow$} & \textbf{28.71} & 0.9029 & 0.1302 & 0.2116

\end{tabular}
}
\caption{\textbf{Comparison of increasing the temporal axis resolution on} \texttt{flame\_salmon} \textbf{scene.} We observe that reducing the normalization range of time does not harm the overall performance. }
\label{tab:supp_resolution}
\end{table}


\section{More Results}
\subsection{Plenoptic Video Dataset}
\label{sup:real_all}

We have demonstrate that our method significantly improves the baseline in the \emph{test view} across all scenes of Unsynchronized Plenoptic Video Dataset.
\fref{fig:supp_real_main} compares the baselines and ours on \texttt{coffee\_martini}, \texttt{cut\_rosted\_beef} and \texttt{sear\_steak} scene. 

We report per-scene test view results in \tref{tab:supp_real_test}. The baselines suffer worse on scenes with larger motion, 
namely \texttt{cook\_spinach} $>$ \texttt{cut\_roasted\_beef} $>$ \texttt{flame\_steak} $>$ \texttt{flame\_salmon} $>$ \texttt{coffee\_martini} $>$ \texttt{sear\_steak}. 
The average \emph{train view} results are in \tref{tab:supp_real_train}. 
Per-scene \emph{test view} results on Synchronized Plenoptic Video Dataset are in \tref{tab:supp_real_sync}.

\begin{figure*}[t]
  \centering
  \includegraphics[width=0.9\linewidth]{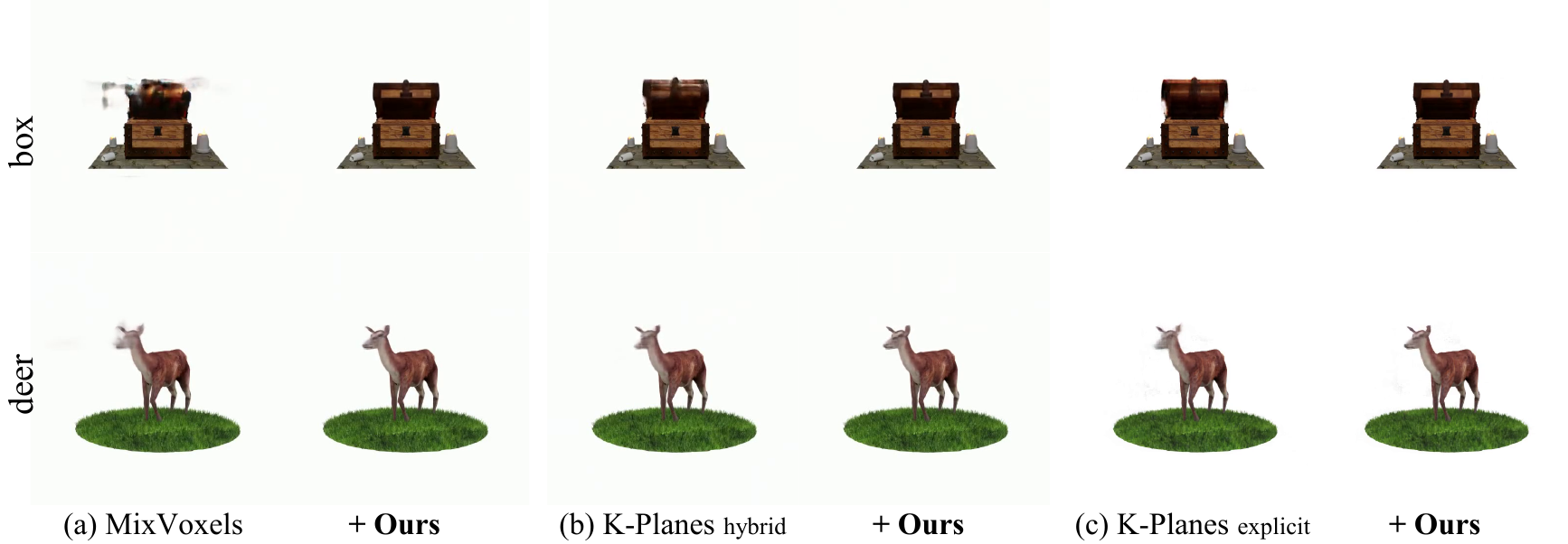}
  \caption{\textbf{More results on Unsynchronized Dynamic Blender Dataset.} Visual comparison of MixVoxels, K-Planes hybrid, K-Planes explicit and our method on them with \texttt{box} and \texttt{dear} scene.}
  \label{fig:supp_blender_qual}
\end{figure*}

\begin{figure*}[t!]
  \centering
  \includegraphics[width=0.9 \linewidth]{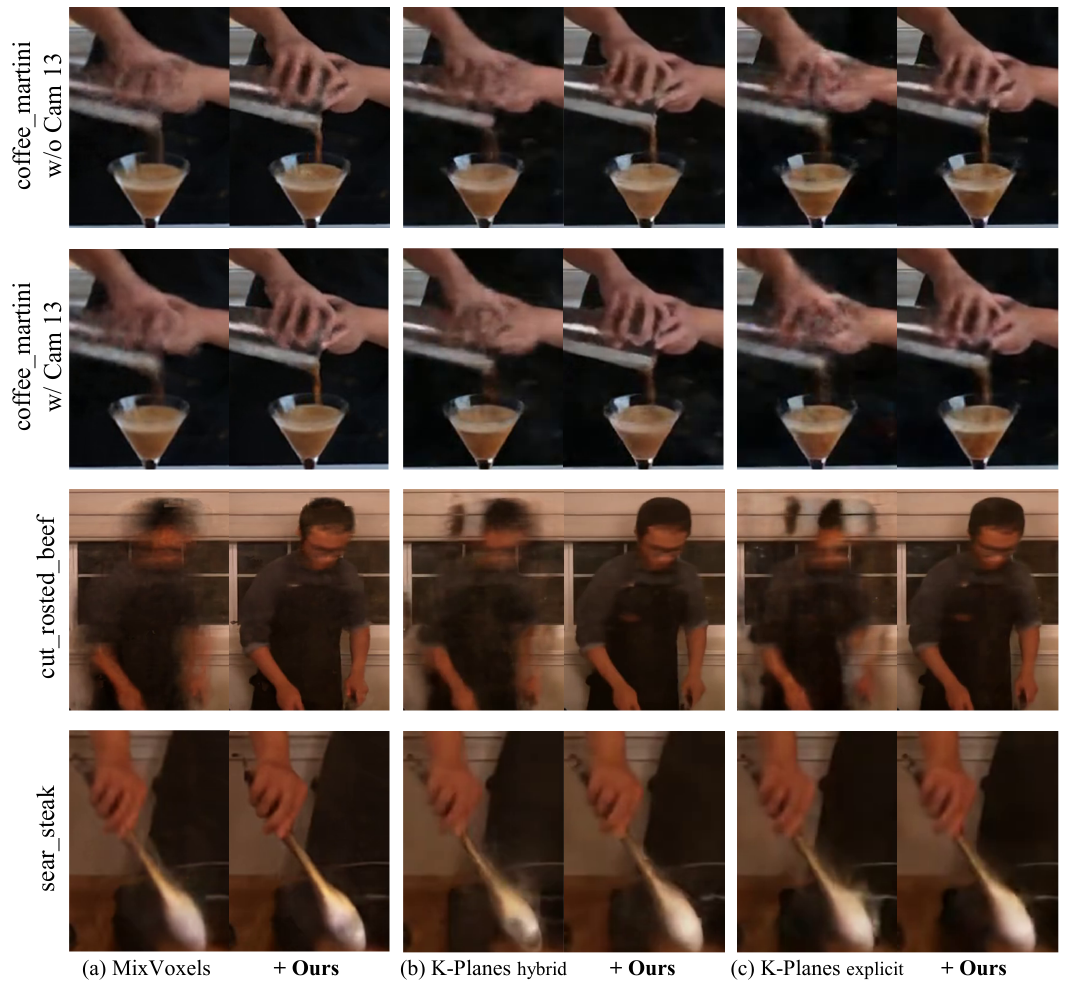}
  \caption{\textbf{More renderings on Unsynchronized Plenoptic Video Dataset.} While the baselines produce severe artifacts, employing our method on them resolves the problem.}
\label{fig:supp_real_main}
\vspace{-2em}
\end{figure*}

    
\subsection{Unsynchronized Dynamic Blender Dataset}
\label{sup:blender_all}
\label{sup:blender_sync}

\fref{fig:supp_blender_qual} compares ours with the baselines on Unsynchronized Dynamic Blender Dataset. On the \texttt{Box} scene, MixVoxels does not properly decompose the dynamic region, leading to some artifacts near the dynamic part on our Sync-MixVoxels. K-Planes surprisingly fails to fit the scene properly, causing deteriorated performance. On the other hand, our method shows a clear reconstruction of the box.

\tref{tab:supp_synthetic_test} reports the per-scene performance on the test views of the Unsynchronized Dynamic Blender Dataset. Our method significantly improves all the baseline, and it is comparable to the results in synchronized setting (\tref{tab:supp_synthetic_sync}).


\begin{table*}[!ht]
\setlength\tabcolsep{3pt}
\renewcommand{\arraystretch}{1}
\centering\resizebox{1.85\columnwidth}{!}{
\begin{tabular}{rl|cccc||rl|cccc}
\multicolumn{2}{c|}{\textbf{test view}} & PSNR & SSIM & \makecell{LPIPS{\scriptsize{alex}}} & \makecell{LPIPS{\scriptsize{vgg}}} & & & PSNR & SSIM & \makecell{LPIPS{\scriptsize{alex}}} & \makecell{LPIPS{\scriptsize{vgg}}} \\ \hline

\multicolumn{6}{c||}{\textbf{coffee\_martini w/o Camera 13}} & 
\multicolumn{6}{c}{\textbf{coffee\_martini w/ Camera 13}} \\ \hline

\multicolumn{1}{r}{MixVoxels} & & 28.75 & 0.8831 & 0.1906 & 0.2752 & 
\multicolumn{1}{r}{MixVoxels} &  & 28.69 & 0.8817 & 0.1932 & 0.2767 \\

\multicolumn{1}{r}{Sync-MixVoxels} & & \textbf{28.96} & \textbf{0.8863} & \textbf{0.1843} & \textbf{0.2697} & \multicolumn{1}{r}{Sync-MixVoxels} & & \textbf{29.06} & \textbf{0.8864} & \textbf{0.1861} & \textbf{0.2716} \\ \hline

K-Planes & \scriptsize{hybrid} & 28.29 & 0.8932 & 0.1446 & 0.2356 & 
K-Planes & \scriptsize{hybrid} & 28.16 & 0.8944 & 0.1438 & 0.2347 \\

Sync-K-Planes & \scriptsize{hybrid} & \textbf{28.34} & \textbf{0.9026} & \textbf{0.1281} & \textbf{0.2158} & 
Sync-K-Planes & \scriptsize{hybrid} & \textbf{28.61} & \textbf{0.9011} & \textbf{0.1343} & \textbf{0.2238} \\ \hline

K-Planes & \scriptsize{explicit} & 27.33 & 0.8807 & 0.1722 & 0.2614 & 
K-Planes & \scriptsize{explicit} & 27.14 & 0.8793 & 0.1763 & 0.2656 \\

Sync-K-Planes & \scriptsize{explicit} & \textbf{28.30} & \textbf{0.9014} & \textbf{0.1369} & \textbf{0.2248} & 
Sync-K-Planes & \scriptsize{explicit} & \textbf{27.88} & \textbf{0.8957} & \textbf{0.1475} & \textbf{0.2361} \\ \hline


\multicolumn{6}{c||}{\textbf{cook\_spinach}} & \multicolumn{6}{c}{\textbf{cut\_rosted\_beef}} \\ \hline

\multicolumn{1}{r}{MixVoxels} &  & 31.28 & 0.9191 & 0.1544 & 0.2659 &
\multicolumn{1}{r}{MixVoxels} &  & 31.24 & 0.9209 & 0.1519 & 0.2633 \\

\multicolumn{1}{r}{Sync-MixVoxels} & & \textbf{31.94} & \textbf{0.9246} & \textbf{0.1381} & \textbf{0.2541} &
\multicolumn{1}{r}{Sync-MixVoxels} & & \textbf{31.67}  & \textbf{0.9253} & \textbf{0.1358} & \textbf{0.2520} \\ \hline 

K-Planes & \scriptsize{hybrid} & 30.05 & 0.9224 & 0.1170 & 0.2179 & 
K-Planes & \scriptsize{hybrid} & 31.07 & 0.9266 & 0.1120 & 0.2124 \\

Sync-K-Planes & \scriptsize{hybrid} & \textbf{32.19} & \textbf{0.9382} & \textbf{0.0890} & \textbf{0.1890} &
Sync-K-Planes  & \scriptsize{hybrid} & \textbf{32.15} & \textbf{0.9361} & \textbf{0.0911} & \textbf{0.1909} \\ \hline

K-Planes & \scriptsize{explicit} & 29.44 & 0.9187 & 0.1302 & 0.2297 &
K-Planes & \scriptsize{explicit} & 30.01 & 0.9185 & 0.1328 & 0.2405 \\

Sync-K-Planes  & \scriptsize{explicit} & \textbf{31.27}  & \textbf{0.9353} & \textbf{0.0981} & \textbf{0.1989} &
Sync-K-Planes  & \scriptsize{explicit} & \textbf{31.56}  & \textbf{0.9326} & \textbf{0.1026} & \textbf{0.2088} \\ \hline


\multicolumn{6}{c||}{\textbf{flame\_salmon}} & \multicolumn{6}{c}{\textbf{flame\_steak}} \\ \hline

\multicolumn{1}{r}{MixVoxels} &  & 27.75 & 0.8750 & 0.2060 & 0.2769 & \multicolumn{1}{r}{MixVoxels} & & 31.14 & 0.9286 & 0.1391 & 0.2508 \\

\multicolumn{1}{r}{Sync-MixVoxels} & & \textbf{28.85} & \textbf{0.8793} & \textbf{0.2022} & \textbf{0.2740} & 
\multicolumn{1}{r}{Sync-MixVoxels} & & \textbf{31.94}  & \textbf{0.9341} & \textbf{0.1266} & \textbf{0.2398} \\ \hline 

K-Planes & \scriptsize{hybrid} & 26.19 & 0.8757 & 0.1698 & 0.2559 & 
K-Planes & \scriptsize{hybrid} & 29.36 & 0.9287 & 0.1146 & 0.2135 \\

Sync-K-Planes & \scriptsize{hybrid} & \textbf{29.17} & \textbf{0.9046} & \textbf{0.1259} & \textbf{0.2112} & 
Sync-K-Planes & \scriptsize{hybrid} & \textbf{31.23} & \textbf{0.9421} & \textbf{0.0897} & \textbf{0.1798} \\ \hline 

K-Planes & \scriptsize{explicit} & 25.61 & 0.8675 & 0.1891 & 0.2728 & 
K-Planes & \scriptsize{explicit} & 29.39 & 0.9251 & 0.1371 & 0.2350 \\

Sync-K-Planes & \scriptsize{explicit} & \textbf{28.56} & \textbf{0.9048} & \textbf{0.1295} & \textbf{0.2114} & 
Sync-K-Planes & \scriptsize{explicit} & \textbf{31.53} & \textbf{0.9453} & \textbf{0.0869} & \textbf{0.1870} \\ \hline
  

\multicolumn{6}{c||}{\textbf{sear\_steak}} & \multicolumn{6}{c}{\textbf{average}} \\ \hline

\multicolumn{1}{r}{MixVoxels} & & 30.88 & 0.9327 & 0.1329 & 0.2453 & 
\multicolumn{1}{r}{MixVoxels} & & 29.96 & 0.9059 & 0.1669 & 0.2648 \\

\multicolumn{1}{r}{Sync-MixVoxels} & & \textbf{31.27} & \textbf{0.9346} & \textbf{0.1259} & \textbf{0.2415} & 
\multicolumn{1}{r}{Sync-MixVoxels} & & \textbf{30.53} & \textbf{0.9101} & \textbf{0.1570} & \textbf{0.2575} \\ \hline 

K-Planes & \scriptsize{hybrid} & 31.03 & 0.9429 & 0.0925 & 0.1854 & 
K-Planes & \scriptsize{hybrid} & 29.16 & 0.9120 & 0.1278 & 0.2222 \\

Sync-K-Planes  & \scriptsize{hybrid} & \textbf{31.40} & \textbf{0.9453} & \textbf{0.0863} & \textbf{0.1815} & 
Sync-K-Planes & \scriptsize{hybrid} & \textbf{30.44} & \textbf{0.9243} & \textbf{0.1064} & \textbf{0.1989} \\ \hline 

K-Planes & \scriptsize{explicit} & 30.66 & 0.9398  & 0.1013 & \textbf{0.2015}        & K-Planes & \scriptsize{explicit} & 28.51 & 0.9042 & 0.1484  & 0.2438 \\

Sync-K-Planes & \scriptsize{explicit} & \textbf{30.71}  & \textbf{0.9407} & \textbf{0.0996} & 0.2049 & Sync-K-Planes & \scriptsize{explicit} & \textbf{29.97} & \textbf{0.9223} & \textbf{0.1144} & \textbf{0.2103} \\ \hline 

\end{tabular}
}
\caption{\textbf{Per-scene performance in the \emph{test view} on Unsynchronized Plenoptic Video Dataset.} Our method improves all the baselines across all scenes.}
\label{tab:supp_real_test}
\end{table*}


\begin{table*}[!ht]
\setlength\tabcolsep{3pt}
\renewcommand{\arraystretch}{1}
\centering\resizebox{1.85\columnwidth}{!}{
\begin{tabular}{rl|cccc||rl|cccc}
\multicolumn{2}{c|}{\textbf{train views}} & PSNR & SSIM & \makecell{LPIPS{\scriptsize{alex}}} & \makecell{LPIPS{\scriptsize{vgg}}} & & & PSNR & SSIM & \makecell{LPIPS{\scriptsize{alex}}} & \makecell{LPIPS{\scriptsize{vgg}}} \\ \hline

\multicolumn{6}{c|}{\textbf{coffee\_martini w/o Camera 13}} & 
\multicolumn{6}{c}{\textbf{coffee\_martini w/ Camera 13}} \\ \hline

\multicolumn{1}{r}{Mixvoxels} & & 29.47           & 0.8875          & 0.1892                  & 0.2733 &
\multicolumn{1}{r}{Mixvoxels} &  & 29.54           & 0.8877          & 0.1909                  & 0.2738 \\

\multicolumn{1}{r}{Sync-Mixvoxels} & & \textbf{29.95}  & \textbf{0.8915} & \textbf{0.1819}         & \textbf{0.2691} & \multicolumn{1}{r}{Sync-Mixvoxels} & & \textbf{30.09}  & \textbf{0.8921} & \textbf{0.1827}         & \textbf{0.2694} \\ \hline

K-Planes & \scriptsize{hybrid} & 29.17           & 0.8926          & 0.1690                  & 0.2544 & 
K-Planes & \scriptsize{hybrid} & 29.42           & 0.8928          & 0.1687                  & 0.2554 \\

Sync-K-Planes & \scriptsize{hybrid} & \textbf{31.09}  & \textbf{0.9138} & \textbf{0.1346}         & \textbf{0.2156}  & 
Sync-K-Planes & \scriptsize{hybrid} & \textbf{29.72}  & \textbf{0.9032} & \textbf{0.1466}         & \textbf{0.2299}  \\ \hline

K-Planes & \scriptsize{explicit} & 29.14           & 0.8921          & 0.1558                  & 0.2465  & 
K-Planes & \scriptsize{explicit} & 27.63           & 0.8625          & 0.2187                  & 0.3000  \\

Sync-K-Planes & \scriptsize{explicit} & \textbf{30.64}  & \textbf{0.9073} & \textbf{0.1334}         & \textbf{0.2230}  & 
Sync-K-Planes & \scriptsize{explicit} & \textbf{29.34}  & \textbf{0.8986} & \textbf{0.1418}         & \textbf{0.2325} \\ \hline


\multicolumn{6}{c||}{\textbf{cook\_spinach}} & \multicolumn{6}{c}{\textbf{cut\_rosted\_beef}} \\ \hline

\multicolumn{1}{r}{Mixvoxels} &  & 32.34           & 0.9253          & 0.1410                  & 0.2498 &
\multicolumn{1}{r}{Mixvoxels} &  & 32.86           & 0.9330          & 0.1289                  & 0.2413  \\

\multicolumn{1}{r}{Sync-Mixvoxels} & & \textbf{33.87}  & \textbf{0.9336} & \textbf{0.1235}         & \textbf{0.2383} &
\multicolumn{1}{r}{Sync-Mixvoxels} & & \textbf{34.43}  & \textbf{0.9398} & \textbf{0.1132}         & \textbf{0.2299} \\ \hline 

K-Planes & \scriptsize{hybrid} & 30.15           & 0.9220          & 0.1247                  & 0.2250 & 
K-Planes & \scriptsize{hybrid} & 29.83           & 0.9200          & 0.1270                  & 0.2302 \\

Sync-K-Planes & \scriptsize{hybrid} & \textbf{30.21}  & \textbf{0.9285} & \textbf{0.1016}         & \textbf{0.1980}       &
Sync-K-Planes  & \scriptsize{hybrid} & \textbf{31.38}  & \textbf{0.9313} & \textbf{0.0994}         & \textbf{0.2000} \\ \hline

K-Planes & \scriptsize{explicit} & 29.34           & 0.9039          & 0.1584                  & 0.2557 &
K-Planes & \scriptsize{explicit} & 30.23           & 0.9231          & 0.1172                  & 0.2162 \\

Sync-K-Planes  & \scriptsize{explicit} & \textbf{30.38}  & \textbf{0.9301} & \textbf{0.0981}         & \textbf{0.1942}  &
Sync-K-Planes  & \scriptsize{explicit} & \textbf{31.92}  & \textbf{0.9347} & \textbf{0.0946}         & \textbf{0.1908} \\ \hline


\multicolumn{6}{c||}{\textbf{flame\_salmon}} & \multicolumn{6}{c}{\textbf{flame\_steak}} \\ \hline

\multicolumn{1}{r}{Mixvoxels} &  & 29.36           & 0.8799          & 0.1979                  & 0.2834               
& \multicolumn{1}{r}{Mixvoxels} &  & \textbf{30.33}           & 0.9292          & 0.1313                  & 0.2430                 \\

\multicolumn{1}{r}{Sync-Mixvoxels} & & \textbf{29.97}  & \textbf{0.8842} & \textbf{0.1899}         & \textbf{0.2789}     & 
\multicolumn{1}{r}{Sync-Mixvoxels} & & 29.94  & \textbf{0.9300} & \textbf{0.1208}         & \textbf{0.2359}         \\ \hline 

K-Planes       & \scriptsize{hybrid}   & 26.98           & 0.8760          & 0.2031                  & 0.2835                   
& K-Planes       & \scriptsize{hybrid}   & 29.33           & 0.9227          & 0.1341                  & 0.2295                  \\

Sync-K-Planes  & \scriptsize{hybrid}   & \textbf{31.07}  & \textbf{0.9131} & \textbf{0.1334}         & \textbf{0.2146}             
& Sync-K-Planes  & \scriptsize{hybrid}   & \textbf{29.83}  & \textbf{0.9370} & \textbf{0.0933}         & \textbf{0.1849}        \\ \hline 

K-Planes       & \scriptsize{explicit} & 27.35           & 0.8792          & 0.1839                  & 0.2779                        
& K-Planes       & \scriptsize{explicit} & 29.58           & 0.9237          & 0.1219                  & 0.2187                 \\

Sync-K-Planes  & \scriptsize{explicit} & \textbf{28.95}  & \textbf{0.9017} & \textbf{0.1400}         & \textbf{0.2312}
& Sync-K-Planes  & \scriptsize{explicit} &  \textbf{30.04}  & \textbf{0.9385} & \textbf{0.0905}         & \textbf{0.1794}                \\ \hline
  

\multicolumn{6}{c||}{\textbf{sear\_steak}} & \multicolumn{6}{c}{\textbf{average}} \\ \hline

\multicolumn{1}{r}{Mixvoxels} &  & 34.04           & 0.9381          & 0.1159                  & 0.2311
& \multicolumn{1}{r}{Mixvoxels} &  & 31.13           & 0.9115          & 0.1565                  & 0.2565                 \\

\multicolumn{1}{r}{Sync-Mixvoxels} &  & \textbf{34.86}  & \textbf{0.9419} & \textbf{0.1076}         & \textbf{0.2250}       
& \multicolumn{1}{r}{Sync-Mixvoxels} & & \textbf{31.87}  & \textbf{0.9162} & \textbf{0.1457}         & \textbf{0.2495}        \\ \hline 

K-Planes       & \scriptsize{hybrid}   & 32.57           & 0.9402          & 0.0963                  & 0.1924                
& K-Planes       & \scriptsize{hybrid}    & 29.64           & 0.9095          & 0.1461                  & 0.2386                 \\

Sync-K-Planes  & \scriptsize{hybrid}   & \textbf{32.24}  & \textbf{0.9398} & \textbf{0.0941}         & \textbf{0.1916}          
& Sync-K-Planes  & \scriptsize{hybrid}   & \textbf{30.79}  & \textbf{0.9238} & \textbf{0.1147}         & \textbf{0.2050}        \\ \hline 

K-Planes       & \scriptsize{explicit} & 31.46           & 0.9244          & 0.1275                  & 0.2189        
& K-Planes       & \scriptsize{explicit} & 29.25           & 0.9013          & 0.1548                  & 0.2477                 \\

Sync-K-Planes  & \scriptsize{explicit} & \textbf{32.87}  & \textbf{0.9435} & \textbf{0.0845}         & \textbf{0.1786}      
& Sync-K-Planes  & \scriptsize{explicit} & \textbf{30.59}  & \textbf{0.9221} & \textbf{0.1118}         & \textbf{0.2042} \\ \hline 

\end{tabular}
}
\caption{\textbf{Per-scene performance in the \emph{train view} on Unsynchronized Plenoptic Video Dataset.} Our method enables all baselines to successfully fit unsynchronized training views across all scenes.}
\label{tab:supp_real_train}
\end{table*}


\begin{table*}[!ht]
\setlength\tabcolsep{3pt}
\centering\resizebox{1.85\columnwidth}{!}{
\begin{tabular}{rl|cccc||rl|cccc}
\multicolumn{2}{c|}{\textbf{test view}} & PSNR & SSIM & \makecell{LPIPS{\scriptsize{alex}}} & \makecell{LPIPS{\scriptsize{vgg}}} & & & PSNR & SSIM & \makecell{LPIPS{\scriptsize{alex}}} & \makecell{LPIPS{\scriptsize{vgg}}} \\ \hline

\multicolumn{6}{c||}{\textbf{coffee\_martini w/o Camera 13}} & 
\multicolumn{6}{c}{\textbf{coffee\_martini w/ Camera 13}} \\ \hline

\multicolumn{1}{r}{MixVoxels}  && \textbf{28.97}  & \textbf{0.8862} & \textbf{0.1840}         & \textbf{0.2681}        
& \multicolumn{1}{r}{MixVoxels}               && 28.79           & 0.8854          & 0.1849                  & 0.2724                 \\

\multicolumn{1}{r}{Sync-MixVoxels} & & 28.76           & 0.8853          & 0.1845                  & 0.2698                 
& \multicolumn{1}{r}{Sync-MixVoxels}        & & \textbf{28.88}  & \textbf{0.8865} & \textbf{0.1817}         & \textbf{0.2713}        \\ \hline

K-Planes & \scriptsize{hybrid} & 28.78           & 0.9034          & \textbf{0.1282}         & \textbf{0.2165}        & K-Planes & \scriptsize{hybrid}        & \textbf{28.80}  & \textbf{0.9025} & \textbf{0.1324}         & \textbf{0.2220}        \\ 

Sync-K-Planes & \scriptsize{hybrid} & \textbf{28.97}  & \textbf{0.9035} & 0.1289                  & 0.2202                
& Sync-K-Planes & \scriptsize{hybrid}   & 28.59           & 0.8994          & 0.1379                  & 0.2260   \\ \hline

K-Planes & \scriptsize{explicit}   & 28.30           & \textbf{0.9017} & \textbf{0.1375}         & \textbf{0.2241}        
& K-Planes & \scriptsize{explicit}      & 27.78           & 0.8946          & 0.1489                  & 0.2384 \\

Sync-K-Planes & \scriptsize{explicit} & \textbf{28.39}  & 0.9016          & 0.1382                  & 0.2254                 
& Sync-K-Planes & \scriptsize{explicit} & \textbf{28.01}  & \textbf{0.8966} & \textbf{0.1464}         & \textbf{0.2351} \\ \hline


\multicolumn{6}{c||}{\textbf{cook\_spinach}} & \multicolumn{6}{c}{\textbf{cut\_rosted\_beef}} \\ \hline

\multicolumn{1}{r}{MixVoxels} &  & \textbf{31.77}  & 0.9247          & 0.1423                  & 0.2578                 
& \multicolumn{1}{r}{MixVoxels}               && \textbf{31.79}  & \textbf{0.9253} & 0.1401                  & 0.2543 \\

\multicolumn{1}{r}{Sync-MixVoxels} & & 31.71           & \textbf{0.9260} & \textbf{0.1399}         & \textbf{0.2538}        
& \multicolumn{1}{r}{Sync-MixVoxels}        & & 31.69           & 0.9249          & \textbf{0.1373}         & \textbf{0.2517} \\ \hline 

K-Planes & \scriptsize{hybrid} & 31.75           & \textbf{0.9389} & \textbf{0.0901}         & 0.1896                 
& K-Planes & \scriptsize{hybrid}        & \textbf{32.26}  & \textbf{0.9375} & \textbf{0.0903}         & \textbf{0.1937} \\

Sync-K-Planes & \scriptsize{hybrid} & \textbf{32.18}  & 0.9385          & 0.0913                  & \textbf{0.1890}        
& Sync-K-Planes & \scriptsize{hybrid}   & 31.82           & 0.9346          & 0.0936                  & 0.1979 \\ \hline

K-Planes & \scriptsize{explicit} & 31.10           & 0.9318          & 0.1037                  & 0.2043                 
& K-Planes & \scriptsize{explicit}      & \textbf{31.91}  & 0.9384          & 0.0963                  & 0.1991 \\

Sync-K-Planes  & \scriptsize{explicit} & \textbf{31.24}  & \textbf{0.9325} & \textbf{0.1030}         & \textbf{0.2034}        
& Sync-K-Planes  & \scriptsize{explicit} & 31.88           & \textbf{0.9386} & \textbf{0.0957}         & \textbf{0.1990} \\ \hline


\multicolumn{6}{c||}{\textbf{flame\_salmon}} & \multicolumn{6}{c}{\textbf{flame\_steak}} \\ \hline

\multicolumn{1}{r}{MixVoxels} &  & 28.79           & 0.8808          & 0.1964                  & 0.2718                 
& \multicolumn{1}{r}{MixVoxels}              & & 31.62           & 0.9326          & 0.1290                  & 0.2440\\

\multicolumn{1}{r}{Sync-MixVoxels} & & \textbf{28.93}  & \textbf{0.8810} & \textbf{0.1952}         & \textbf{0.2692}        
& \multicolumn{1}{r}{Sync-MixVoxels}         & & \textbf{31.68}  & \textbf{0.9341} & \textbf{0.1263}         & \textbf{0.2386} \\ \hline 

K-Planes & \scriptsize{hybrid} & 28.98           & 0.9029          & 0.1285                  & 0.2152                 
& K-Planes & \scriptsize{hybrid}        & 31.07           & \textbf{0.9477} & \textbf{0.0814}         & 0.1754 \\

Sync-K-Planes & \scriptsize{hybrid} & \textbf{29.15}  & \textbf{0.9032} & \textbf{0.1275}         & \textbf{0.2138}        
& Sync-K-Planes & \scriptsize{hybrid}   & \textbf{31.51}  & 0.9458          & 0.0830                  & \textbf{0.1745} \\ \hline 

K-Planes & \scriptsize{explicit} & 28.63           & 0.9034          & 0.1303                  & 0.2139                 
& K-Planes & \scriptsize{explicit}      & 31.30           & 0.9446          & 0.0894                  & 0.1908 \\

Sync-K-Planes & \scriptsize{explicit} & \textbf{28.79}  & \textbf{0.9051} & \textbf{0.1277}         & \textbf{0.2114}        
& Sync-K-Planes & \scriptsize{explicit} & \textbf{31.47}  & \textbf{0.9451} & \textbf{0.0871}         & \textbf{0.1870} \\ \hline
  

\multicolumn{6}{c||}{\textbf{sear\_steak}} & \multicolumn{6}{c}{\textbf{average}} \\ \hline

\multicolumn{1}{r}{MixVoxels} & & 30.98           & \textbf{0.9350} & 0.1271                  & 0.2419                 
& \multicolumn{1}{r}{MixVoxels}              & & 30.39           & 0.9100          & 0.1577                  & 0.2586 \\

\multicolumn{1}{r}{Sync-MixVoxels} & & \textbf{31.24}  & 0.9349          & \textbf{0.1262}         & \textbf{0.2406}        
& \multicolumn{1}{r}{Sync-MixVoxels}         & & \textbf{30.41}  & \textbf{0.9104} & \textbf{0.1559}         & \textbf{0.2564} \\ \hline 

K-Planes & \scriptsize{hybrid} & 31.16           & 0.9471          & 0.0801                  & 0.1739                 
& K-Planes & \scriptsize{hybrid}        & 30.40           & \textbf{0.9257}          & \textbf{0.1044}         & \textbf{0.1980} \\

Sync-K-Planes  & \scriptsize{hybrid} & \textbf{31.70}  & \textbf{0.9474} & \textbf{0.0795}         & \textbf{0.1771}        
& Sync-K-Planes  & \scriptsize{hybrid}   & \textbf{30.56}  & 0.9246 & 0.1059                  & 0.1998 \\ \hline 

K-Planes & \scriptsize{explicit} & \textbf{31.24}  & 0.9458          & \textbf{0.0855}         & \textbf{0.1874}        
& K-Planes & \scriptsize{explicit}      & 30.04           & 0.9229          & 0.1131                  & 0.2083 \\

Sync-K-Planes & \scriptsize{explicit} & 31.18           & \textbf{0.9466} & 0.0866                  & 0.1893                 
& Sync-K-Planes & \scriptsize{explicit} & \textbf{30.14}  & \textbf{0.9237} & \textbf{0.1121}         & \textbf{0.2072} \\ \hline 

\end{tabular}
}
\caption{\textbf{Per-scene performance in the \emph{test view} on Synchronized Plenoptic Video Dataset.} Our method improves the baselines even on the synchronized setting by resolving the small temporal errors.}
\label{tab:supp_real_sync}
\end{table*}


\begin{table*}[!ht]
\setlength\tabcolsep{3pt}
\renewcommand{\arraystretch}{1}
\centering\resizebox{1.95\columnwidth}{!}{
\begin{tabular}{rl|cccc||rl|cccc}
\multicolumn{2}{c|}{\textbf{test view}} & PSNR & SSIM & \makecell{LPIPS{\scriptsize{alex}}} & \makecell{LPIPS{\scriptsize{vgg}}} & & & PSNR & SSIM & \makecell{LPIPS{\scriptsize{alex}}} & \makecell{LPIPS{\scriptsize{vgg}}} \\ \hline

\multicolumn{6}{c||}{\textbf{box}} & 
\multicolumn{6}{c}{\textbf{deer}} \\ \hline

\multicolumn{1}{r}{MixVoxels} & & 29.08& 0.9801& 0.018& 0.021& \multicolumn{1}{r}{MixVoxels} & & 31.48& 0.9635& 0.031& 0.049\\

\multicolumn{1}{r}{Sync-MixVoxels} & & \textbf{39.29}& \textbf{0.9927}& \textbf{0.007}& \textbf{0.013}& \multicolumn{1}{r}{Sync-MixVoxels}         & & 
\textbf{33.71}& \textbf{0.9682}& \textbf{0.022}& \textbf{0.039}\\ \hline

K-Planes & \scriptsize{hybrid} & 31.31& 0.9793& 0.018& 0.023& K-Planes & \scriptsize{hybrid}        & 33.14& 0.9671& 0.017& 0.027\\

Sync-K-Planes & \scriptsize{hybrid} & \textbf{39.99}& \textbf{0.9922}& \textbf{0.007}& \textbf{0.011}& Sync-K-Planes & \scriptsize{hybrid}   & \textbf{36.40}& \textbf{0.9718}& \textbf{0.009}& \textbf{0.019}\\ \hline

K-Planes & \scriptsize{explicit} & 27.63& 0.9712& 0.039& 0.047& K-Planes & \scriptsize{explicit}      & 33.87& 0.9683& 0.018& 0.030\\

Sync-K-Planes & \scriptsize{explicit} & \textbf{39.76}& \textbf{0.9920}& \textbf{0.007}& \textbf{0.012}& Sync-K-Planes & \scriptsize{explicit} & \textbf{36.49}& \textbf{0.9732}& \textbf{0.010}& \textbf{0.019}\\ \hline


\multicolumn{6}{c||}{\textbf{fox}} & \multicolumn{6}{c}{\textbf{average}} \\ \hline

\multicolumn{1}{r}{MixVoxels} &  & 32.61& 0.9822& 0.022& 0.031& \multicolumn{1}{r}{MixVoxels}             & & 31.06& 0.9753& 0.024& 0.034\\

\multicolumn{1}{r}{Sync-MixVoxels} & & \textbf{38.33}& \textbf{0.9915}& \textbf{0.007}& \textbf{0.017}& \multicolumn{1}{r}{Sync-MixVoxels}         & & \textbf{37.11}& \textbf{0.9841}& \textbf{0.012}& \textbf{0.023}\\ \hline 

K-Planes & \scriptsize{hybrid} & 33.54& 0.9850& 0.018& 0.023& K-Planes & \scriptsize{hybrid}        & 32.66& 0.9771& 0.017& 0.024\\

Sync-K-Planes & \scriptsize{hybrid} & \textbf{41.82}& \textbf{0.9947}& \textbf{0.004}& \textbf{0.009}& Sync-K-Planes & \scriptsize{hybrid}   & \textbf{39.40}& \textbf{0.9863}& \textbf{0.007}& \textbf{0.013}\\ \hline

K-Planes & \scriptsize{explicit} & 34.94& 0.9857& 0.017& 0.024& K-Planes & \scriptsize{explicit}      & 32.15& 0.9751& 0.025& 0.034\\

Sync-K-Planes  & \scriptsize{explicit} & \textbf{41.81}& \textbf{0.9943}& \textbf{0.004}& \textbf{0.011}& Sync-K-Planes  & \scriptsize{explicit} & \textbf{39.35}& \textbf{0.9865}& \textbf{0.007}& \textbf{0.014}\\ \hline

\end{tabular}
}
\caption{\textbf{Per-scene performance in the \emph{test view} on Unsynchronized Dynamic Blender Dataset.} Our method enhances performance on all the baselines.}
\label{tab:supp_synthetic_test}
\end{table*}
\textbf{}


\begin{table*}[!ht]
\setlength\tabcolsep{3pt}
\renewcommand{\arraystretch}{1}
\centering\resizebox{1.95\columnwidth}{!}{
\begin{tabular}{rl|cccc||rl|cccc}
\multicolumn{2}{c|}{\textbf{test view}} & PSNR & SSIM & \makecell{LPIPS{\scriptsize{alex}}} & \makecell{LPIPS{\scriptsize{vgg}}} & & & PSNR & SSIM & \makecell{LPIPS{\scriptsize{alex}}} & \makecell{LPIPS{\scriptsize{vgg}}} \\ \hline

\multicolumn{6}{c||}{\textbf{box}} & 
\multicolumn{6}{c}{\textbf{deer}} \\ \hline

\multicolumn{1}{r}{MixVoxels} & & \textbf{40.36}& \textbf{0.9941}& \textbf{0.005}& \textbf{0.010}& \multicolumn{1}{r}{MixVoxels}              & & \textbf{33.72}& 0.9682& \textbf{0.022}& 0.040\\

\multicolumn{1}{r}{Sync-MixVoxels} & & 39.80& 0.9933& 0.006& 0.011& \multicolumn{1}{r}{Sync-MixVoxels}         & & 33.70& \textbf{0.9692}& \textbf{0.022}& \textbf{0.038}\\ \hline

K-Planes & \scriptsize{hybrid} & \textbf{40.14}& \textbf{0.9923}& \textbf{0.006}& \textbf{0.011}& K-Planes & \scriptsize{hybrid}        & \textbf{36.93}& \textbf{0.9732}& \textbf{0.007}& 0.018\\

Sync-K-Planes & \scriptsize{hybrid} & 40.06& 0.9922& \textbf{0.006}& \textbf{0.011}& Sync-K-Planes & \scriptsize{hybrid}   & 36.87& 0.9723& \textbf{0.007}& \textbf{0.017}\\ \hline

K-Planes & \scriptsize{explicit}& 39.85& 0.9919& \textbf{0.007}& \textbf{0.012}& K-Planes & \scriptsize{explicit}      & \textbf{37.08}& \textbf{0.9742}& \textbf{0.007}& \textbf{0.017}\\

Sync-K-Planes & \scriptsize{explicit} & \textbf{39.86}& \textbf{0.9921}& \textbf{0.007}& \textbf{0.012}& Sync-K-Planes & \scriptsize{explicit} & 36.99& 0.9739& 0.008& 0.018\\ \hline


\multicolumn{6}{c||}{\textbf{fox}} & \multicolumn{6}{c}{\textbf{average}} \\ \hline

\multicolumn{1}{r}{MixVoxels} &  & 38.64& 0.9918& \textbf{0.007}& \textbf{0.015}& \multicolumn{1}{r}{MixVoxels}              & & \textbf{37.57}& 0.9847& \textbf{0.011}& 0.022\\

\multicolumn{1}{r}{Sync-MixVoxels} & & \textbf{38.85}&\textbf{0.9921}& \textbf{0.007}& \textbf{0.015}& \multicolumn{1}{r}{Sync-MixVoxels}         & & 37.45& \textbf{0.9849}& 0.012& \textbf{0.021}\\ \hline 

K-Planes & \scriptsize{hybrid} & 42.49& 0.9951& \textbf{0.003}& \textbf{0.009}& K-Planes & \scriptsize{hybrid}        & \textbf{39.85}& \textbf{0.9868}& \textbf{0.006}& \textbf{0.012}\\

Sync-K-Planes & \scriptsize{hybrid} & \textbf{42.52}& \textbf{0.9952}& \textbf{0.003}& \textbf{0.009}& Sync-K-Planes & \scriptsize{hybrid}   & 39.82& 0.9866& \textbf{0.006}& \textbf{0.012}\\ \hline

K-Planes & \scriptsize{explicit} & 42.09& 0.9946& \textbf{0.003}& \textbf{0.009}& K-Planes & \scriptsize{explicit}      & \textbf{39.67}& \textbf{0.9869}& \textbf{0.006}& \textbf{0.013}\\

Sync-K-Planes  & \scriptsize{explicit} & \textbf{42.14}& \textbf{0.9947}& 0.004& \textbf{0.009}& Sync-K-Planes  & \scriptsize{explicit} & \textbf{39.67}& \textbf{0.9869}& \textbf{0.006}& \textbf{0.013}\\ \hline

\end{tabular}
}
\caption{\textbf{Per-scene performance in the \emph{test view} on Synchronized Dynamic Blender Dataset.} The results of our method in the unsynchronized setting (\tref{tab:supp_synthetic_test}) are nearly comparable to the performance in the synchronized setting. } 
\label{tab:supp_synthetic_sync}
\end{table*}


\end{appendix}

\end{document}